%% file: main.tex
\documentclass[fleqn,10pt]{wlscirep}
\input{inc/packages}

\title{Machine Theory of Mind for Autonomous Cyber-Defence}

\author[1,*]{Luke Swaby}
\author[1]{Matthew Stewart}
\author[1]{Daniel Harrold}
\author[1]{Chris Willis}
\author[1]{Gregory Palmer}
\affil[1]{BAE Systems Applied Intelligence Labs,
Chelmsford Office \& Technology Park,
Chelmsford,
Essex,
CM2 8HN.}

\affil[*]{luke.swaby2@baesystems.com}

\input{inc/newcommands}

\begin{abstract}
Intelligent autonomous agents hold much potential for the domain of cyber security. However, due to many state-of-the-art approaches relying on uninterpretable black-box models, there is growing demand for methods that offer stakeholders clear and actionable insights into their latent beliefs and motivations. To address this, we evaluate Theory of Mind (ToM) approaches for Autonomous Cyber Operations. Upon learning a robust prior, ToM models can predict an agent's goals, behaviours, and contextual beliefs given only a handful of past behaviour observations. In this paper, we introduce a novel Graph Neural Network (GNN)-based ToM architecture tailored for cyber-defence, Graph-In, Graph-Out (GIGO)-ToM, which can accurately predict both the targets and attack trajectories of adversarial cyber agents over arbitrary computer network topologies. To evaluate the latter, we propose a novel extension of the Wasserstein distance for measuring the similarity of graph-based probability distributions. Whereas the standard Wasserstein distance lacks a fixed reference scale, we introduce a graph-theoretic normalization factor that enables a standardized comparison between networks of different sizes. We furnish this metric, which we term the \emph{Network Transport Distance} (NTD), with a weighting function that emphasizes predictions according to custom node features, allowing network operators to explore arbitrary strategic considerations. Benchmarked against a Graph-In, Dense-Out (GIDO)-ToM architecture in an abstract cyber-defence environment, our empirical evaluations show that GIGO-ToM can accurately predict the goals and behaviours of various unseen cyber-attacking agents across a range of network topologies, as well as learn embeddings that can effectively characterize their policies.

\end{abstract}

\begin{document}

\flushbottom
\maketitle
%
%
\thispagestyle{firstpage}


\input{sections/introduction}

\input{sections/contributions}

\input{sections/related_work}

\input{sections/preliminaries}
\input{sections/methods}
\input{sections/results}

\input{sections/discussion}

\bibliography{sample}






\section*{Acknowledgements}

Research funded by Frazer-Nash Consultancy Ltd. on behalf of the Defence Science
and Technology Laboratory (Dstl) which is an executive agency of the UK Ministry
of Defence providing world class expertise and delivering cutting-edge science
and technology for the benefit of the nation and allies. The research supports the
Autonomous Resilient Cyber Defence (ARCD) project within the Dstl Cyber Defence
Enhancement programme.

\section*{Author contributions statement}


All authors provided critical feedback and helped shape the research.
L.S and G.P contributed to the development and evaluation of the
\gido and \gigo approaches, along with the experiment design.
L.S formulated the \mtrclong for evaluating
successor representations within the context of cyber-defence
scenarios.
D.H. was responsible for implementing the graph-based observation wrapper
for YAWNING-Titan.
D.H. and M.S were responsible for implementing rules-based cyber-attack and
cyber-defence agents for data gathering and evaluation.
C.W. provided an extensive technical review.
G.P. coordinated and supervised the project.

\section*{Data availability}

The framework and data that support the findings of this study are proprietary to
BAE Systems and restrictions apply to the availability of these items which are not
publicly available.
Request for access to these items can be sent to Luke Swaby (\href{luke.swaby2@baesystems.com}{luke.swaby2@baesystems.com})
and Gregory Palmer (\href{gregory.palmer@baesystems.com}{gregory.palmer@baesystems.com}).
Requests will be subject to BAE Systems standard data protection policies and processes.

\appendix
\section*{Appendices}
\addcontentsline{toc}{section}{Appendices}
\renewcommand{\thesubsection}{\Alph{subsection}}

\input{appendix/networks.tex}

\input{appendix/agents.tex}

\input{appendix/ntd_loss.tex}

\end{document}

%% file: inc/packages.tex
\usepackage[utf8]{inputenc}
\usepackage[T1]{fontenc}
\usepackage{float,subfig,graphicx}
\usepackage{enumitem}
\usepackage{hyperref}

\usepackage{color,soul}
\usepackage{amsmath}

\DeclareMathAlphabet{\mathcal}{OMS}{cmsy}{m}{n}

%% file: inc/newcommands.tex
\input{inc/colors}
\usepackage{amsfonts}
\usepackage{amsmath}
\usepackage{amsthm}
\usepackage{xspace}
\usepackage{bm}

\newcommand{\comLS}[1]{\textcolor{red}{\textbf{\large [}\colorbox{yellow}{\textbf{Luke:}}{\small #1}\textbf{\large ]}}}

\newcommand{\M}{\mathcal{M}}

\newcommand{\obs}{(obs)}

\newcommand{\SR}{\hat{SR}}
\newcommand{\TargetNode}{\hat{v}}

\newcommand{\hvn}{high-value node\xspace}
\newcommand{\hvns}{high-value nodes\xspace}
\newcommand{\hvt}{high-value target\xspace}

\newcommand{\hvus}{high-value users\xspace}
\newcommand{\agents}{\mathcal{N}}

\newcommand{\states}{\mathcal{S}}
\newcommand{\state}{s}
\newcommand{\stateobs}{\state_{t}^{\obs}}
\newcommand{\stateobsfun}{\omega^{\obs}}

\newcommand{\actions}{\mathcal{A}}
\newcommand{\action}{a}
\newcommand{\actionobs}{\action_{t}^{\obs}}

\newcommand{\actionobsfun}{\alpha^{\obs}}

\newcommand{\rewards}{\mathcal{R}}
\newcommand{\transition}{\mathcal{P}}
\newcommand{\observations}{\Omega}
\newcommand{\observationf}{\mathcal{O}}
\newcommand{\observation}{o}

\newcommand{\red}{Red\xspace}
\newcommand{\blue}{Blue\xspace}

\newcommand{\eg}{e.g. }
\newcommand{\ie}{i.e. }

\newcommand{\commentout}[1]{}

\newcommand{\rats}{\mathbb{R}}

 \providecommand{\argsA}[2]{ {#1}_{#2} } 

\global\long\def\O#1#2{\argsA{O}{#1}(#2)}

\newcommand{\yt}{YAWNING-TITAN\xspace}
\newcommand{\tom}{ToM\xspace}
\newcommand{\mtom}{MToM\xspace}
\newcommand{\tomfull}{Theory of Mind\xspace}
\newcommand{\mtomfull}{Machine Theory of Mind\xspace}
\newcommand{\tomnet}{ToMnet\xspace}

\newcommand{\gigo}{GIGO-ToM\xspace}
\newcommand{\gido}{GIDO-ToM\xspace}

\newcommand{\character}{e_{c}}
\newcommand{\characterfull}{e_{char}}
\newcommand{\characteri}{e_{c,i}}
\newcommand{\mental}{e_{m}}
\newcommand{\mentalfull}{e_{mental}}
\newcommand{\mentali}{e_{m,i}}

\newcommand{\Lhvti}{\mathcal{L}_{hvt, i}}

\newcommand{\Lsri}{\mathcal{L}_{{sr}, i}}

\newcommand{\Lhvt}{\mathcal{L}_{hvt}}

\newcommand{\Lsr}{\mathcal{L}_{sr}}
\newcommand{\ntdloss}{\text{NTD}_{\mathcal{L}}}

\providecommand{\Bigbrackets}[1]{\Big(#1\Big)}
\providecommand{\bigbrackets}[1]{\big(#1\big)}
\newcommand{\tauobs}{\tau^{\obs}}

\newcommand{\tree}{\emph{TreeNetwork}\xspace}
\newcommand{\treeThirty}{\emph{TreeNetwork30}\xspace}
\newcommand{\treeForty}{\emph{TreeNetwork40}\xspace}
\newcommand{\treeFifty}{\emph{TreeNetwork50}\xspace}
\newcommand{\treeSeventy}{\emph{TreeNetwork70}\xspace}
\newcommand{\treeNinety}{\emph{TreeNetwork90}\xspace}
\newcommand{\treeMixed}{\emph{TreeNetworkMixed}\xspace}

\newcommand{\oc}{\emph{OpticalCoreNetwork}\xspace}
\newcommand{\forest}{\emph{ForestNetwork}\xspace}

\newcommand{\blueSleep}{\texttt{BlueSleep}\xspace}
\newcommand{\blueRandom}{\texttt{BlueRandom}\xspace}
\newcommand{\blueRandomSmart}{\texttt{BlueRandomSmart}\xspace}
\newcommand{\blueIsolate}{\texttt{BlueIsolate}\xspace}
\newcommand{\blueMSND}{\texttt{BlueMSN-D}\xspace}
\newcommand{\blueMSNS}{\texttt{BlueMSN-S}\xspace}
\newcommand{\blueMSNRNV}{\texttt{BlueMSN-RNV}\xspace}
\newcommand{\blueMSNRestore}{\texttt{BlueMSN-Restore}\xspace}
\newcommand{\blueMSNRNVRestore}{\texttt{BlueMSN-RNV-Restore}\xspace}
\newcommand{\blueRestore}{\texttt{BlueRestore}\xspace}
\newcommand{\redRandomSimple}{\texttt{RedRandomSimple}\xspace}
\newcommand{\redRandomSmart}{\texttt{RedRandomSmart}\xspace}
\newcommand{\redHVTSimple}{\texttt{RedHVTSimple}\xspace}
\newcommand{\redHVTPreference}{\texttt{RedHVTPreference}\xspace}

\newcommand{\redHVTPreferenceSP}{\texttt{RedHVTPreferenceSP}\xspace}
\newcommand{\redTargetConnected}{\texttt{RedTargetConnected}\xspace}
\newcommand{\redTargetUnconnected}{\texttt{RedTargetUnconnected}\xspace}
\newcommand{\redTargetVulnerable}{\texttt{RedTargetVulnerable}\xspace}
\newcommand{\redTargetResilient}{\texttt{RedTargetResilient}\xspace}

\newcommand{\donothing}{\texttt{DoNothing}\xspace}
\newcommand{\scan}{\texttt{Scan}\xspace}
\newcommand{\msn}{\texttt{MakeSafeNode}\xspace}
\newcommand{\rnv}{\texttt{ReduceNodeVulnerability}\xspace}
\newcommand{\restore}{\texttt{Restore}\xspace}
\newcommand{\isolate}{\texttt{Isolate}\xspace}
\newcommand{\reconnect}{\texttt{Reconnect}\xspace}
\newcommand{\basicattack}{\texttt{BasicAttack}\xspace}
\newcommand{\randommove}{\texttt{RandomMove}\xspace}
\newcommand{\zda}{\texttt{ZeroDayAttack}\xspace}
\newcommand{\spread}{\texttt{Spread}\xspace}
\newcommand{\intrude}{\texttt{Intrude}\xspace}
\newcommand{\hdu}{hot-desking user problem\xspace}

\newcommand{\srsq}{\text{NTD}_{\theta}}
\newcommand{\mtrc}{NTD\xspace}
\newcommand{\mtrclong}{Network Transport Distance\xspace}
\newcommand{\Wfn}{\mathcal{W}}
\newcommand{\Wfnfull}{\Wfn: \mathbb{R}^{n \times m} \times [-1, 1]^m \times [0, 1] \to [0, 1]^n}

\newcommand{\w}{\bar{w}}

\newcommand{\graph}{\mathcal{G}}
\newcommand{\feats}{\mathcal{X}}
\newcommand{\feat}[1]{\bar{x}_{#1}}
\newcommand{\featsw}{\mathcal{C}}
\newcommand{\featw}[1]{c_{#1}}
\newcommand{\minmax}[2]{\left\| #1 \right\|_{#2}}


%% file: inc/colors.tex
\definecolor{black}{rgb}		{0.0, 0.0, 0.0}
\definecolor{white}{rgb}		{1.0, 1.0, 1.0}
\definecolor{yellow}{rgb}		{1.0, 1.0, 0.8}
\definecolor{red}{rgb}			{0.6, 0.0, 0.2}
\definecolor{green}{rgb}		{0.6, 0.8, 0.8}
\definecolor{dark_green}{RGB} {0, 140, 0}
\definecolor{gold}{rgb}		{0.6, 0.4, 0.1}
\definecolor{grey}{RGB}{0,0,0}
\definecolor{Gray}{gray}{0.8}
\definecolor{MediumGray}{gray}{0.9}
\definecolor{LightGray}{gray}{0.98}
\definecolor{LightCyan}{rgb}{0.88,1,1}
\definecolor{purple}{RGB}{128,0,128}
\definecolor{sl_blue}{RGB}{47, 60, 105}
\definecolor{orange}{RGB}{255,165,0}
\definecolor{Gray}{gray}{0.85}

%% file: sections/introduction.tex
\section*{Introduction}

As autonomous systems are entrusted with increasingly critical tasks across various industries, the need for explainability has become paramount \cite{pendleton2017perception,scharre2018army,yang2017medical,benitti2012exploring,preece2018stakeholders,Arulkumaran_2017}. Recent advancements have established deep neural networks (DNNs) as the cornerstone of autonomous systems due to their proficiency in handling the complex, high-dimensional data required to scale to real-world environments \cite{tang2022perception,deng2021deep,adawadkar2022cyber,li2019reinforcement}. However, in safety-critical domains where decisions can have far-reaching consequences, the opacity of these `black-box' models poses significant risks, raising concerns around trust, accountability, and safety \cite{rudin2019stop,du2019techniques}. One such domain is that of cybersecurity, where the proliferation of complex computer networks with ever-growing attack surfaces has fueled a cyber arms race with both attackers and defenders increasingly leveraging DNNs for their respective strategies \cite{palmer2023deep,miles2024reinforcement,deng2020edge,ncsc2024,nisioti2018intrusion,kala2023critical,police2020acsc}. Consequently, there is growing urgency for methods to interpret and analyze the latent decision-making processes of these models.

In recent years, a plethora of approaches have emerged aimed at demystifying DNNs. In the context of cybersecurity, these include Explainable Artificial Intelligence (XAI) techniques for malware and anomaly detection \cite{moustafa2023explainable}, as well as solutions that rationalize the actions of deep reinforcement learning agents \cite{greydanus2018visualizing,burke2020robust}. However, these approaches typically focus on post-hoc, data-centric clarifications of \emph{how} a model arrived at a particular decision rather than \emph{why}, and thus lack the level of contextual detail that humans generally deem important when making high-stakes decisions \cite{antoniadi2021current,miller2019explanation,rudin2019stop,lipton2018mythos,miller2019explanation,samek2019explainable}. This limitation becomes especially pronounced in multi-agent decision-making scenarios where understanding the intentions, strategies, and circumstances of other entities can significantly impact outcomes. For example, a computer network defender capable only of identifying the presence and mechanics of a given attack is at a significant disadvantage compared to one who can additionally infer the adversary's objectives. Without understanding \emph{why} certain actions occur, the defensive strategies of the former risk being reactive and insufficient in anticipating evolving threats.


In the psychological literature, the human ability to infer the latent mental states of others is known as a \textit{`\tomfull'} (\tom) \cite{premack1978does}.
An extensive corpus of cognitive studies has linked the absence of ToM with significant social deficiencies in humans \cite{baron1985does,garfield2001social,hamilton2009research,ahmed2011executive,nichols2003mindreading}. This has lead several researchers to investigate whether the intersubjective capabilities of AI systems can be improved with \tom-inspired  model architectures.
This topic was first explored in \emph{\mtomfull} (\mtom) by Rabinowitz et al.\cite{rabinowitz2018machine}, who implemented an observing agent---a parameterized model, dubbed \emph{`\tomnet'}---that employs meta-learning to make predictions over the goals, strategies, and characters of observed agents based solely on past behaviour observations.

\begin{figure}[H]
    \centering
    \includegraphics[width=0.6\columnwidth]{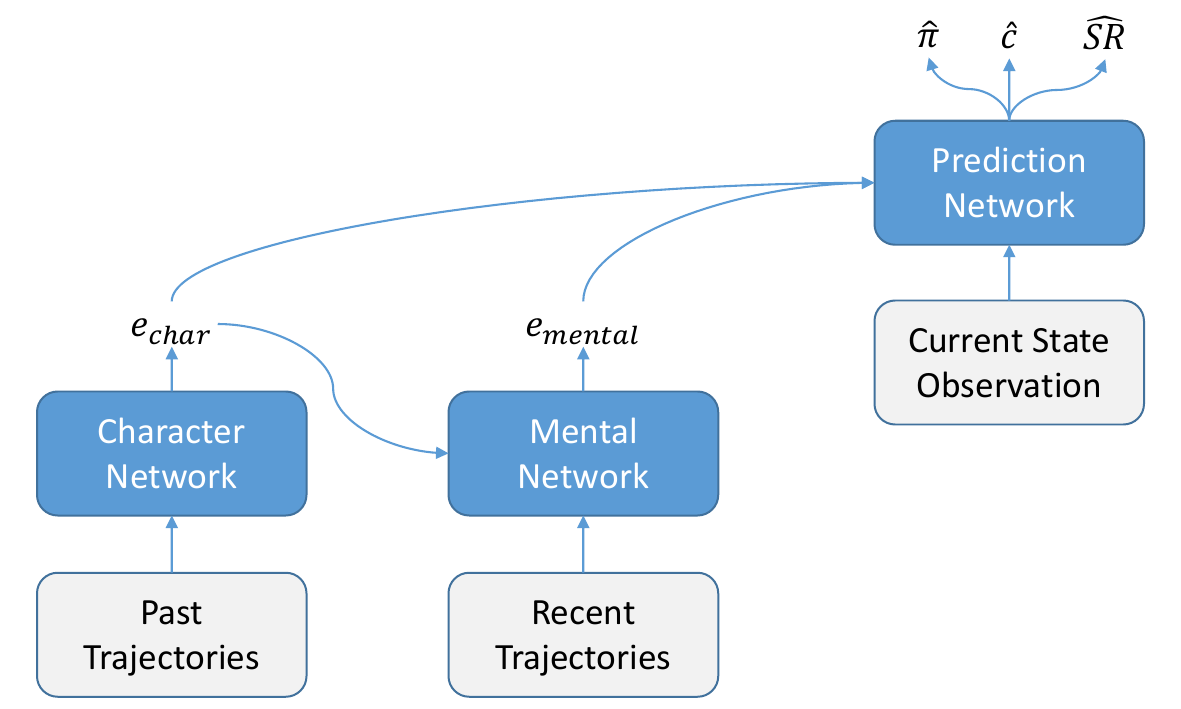}
    \caption{The original \tomnet architecture, adopted from Rabinowitz et al. \cite{rabinowitz2018machine}.
    It consists of three components: a \emph{character network},
    a \emph{mental network},
    and a \emph{prediction network}.
    The character network parses observations of past behaviour
    (e.g. previous episodes) to build a character embedding $\characterfull$.
    The mental network then uses this along with behavioural observations from
    a current episode to build a representation of the agent's current beliefs
    and intentions: the mental state embedding $\mentalfull$.
    Finally, the predictor network utilizes both these embeddings along with a
    snapshot of the current state of the environment to forecast future behaviour:
    the next-step action probabilities $\hat{\pi}$,
    probabilities of whether certain objects will be consumed $\hat{c}$,
    and the predicted successor representations $\SR$ \cite{dayan1993improving}.}
    \label{fig:ToM}
\end{figure}

The \tomnet architecture (\autoref{fig:ToM}) features 3 submodules that respectively parse:  i) a set of past behavior observations; ii) behavior observations from a current episode up to a timestep $t$; and iii) an observation of the state of the environment at $t$. The outputs of each module are propagated forward to a unified prediction mechanism that predicts next-step action probabilities $\hat{\pi}$, the object that the observed agent is targeting $\hat{c}$, and the expected state occupancy, or, \emph{successor representation}, $\SR$ \cite{dayan1993improving}.
Trained and evaluated in simple gridworld environments, Rabinowitz et al.\cite{rabinowitz2018machine} found that \tomnet was able to characterize the behaviours and intentions of a broad taxonomy of agents, and showed strong generalization capabilities.

Intuitively, \tomnet offers a clear and effective paradigm for understanding the decisions made by agents in a cybersecurity context. However, whilst it has been applied to specialised cyber-defence problems before (see related work), our review of existing literature found little work on developing general, network-agnistic, and scalable \tomnet formulations for this domain. As such, our aim in this paper is to provide an abstract evaluation of whether \tomnet can be effectively utilized for cyber-defence.

Right away we can identify several limitations in \tomnet's original formulation that present challenges when applying it to cyber-defence. First, in the original paper, the gridworld environment dimensions were fixed at $11 \times 11$ for all experiments, allowing \tomnet's input and output layers to be fixed accordingly \cite{rabinowitz2018machine}. In cyber-defence scenarios, by contrast, the number of nodes in a network is not necessarily known \emph{a priori}. The \tomnet architecture must therefore be adapted to handle variable sized inputs in order to qualify as a viable candidate for cyber applications. Furthermore, gridworld environments contain relatively few cells that lack individual features, and agents that can only perform a fixed, limited set of actions. Cyber-defence environments, in contrast, are typically dynamic, with significantly larger and more heterogeneous sets of nodes, states, actions, and features. This raises the question of whether \tomnet can scale to environments suffering the `curse-of-dimensionality'.

The domain-specific nature of our experiments also necessitates bespoke evaluation methodologies. For example, in the original paper, predicted successor representations were qualitatively evaluated against agents' true behaviours \cite{rabinowitz2018machine}. The key strength of this approach is that it enables an intuitive understanding of precitive idiosyncrasies and the extent to which the model can approximate the nuances of human-like reasoning (in keeping with the exploratory nature of the study). However, our aims here are slightly different in that we intend to quantitively evaluate \tomnet's performance across various experimental settings in order to determine its viability for real-world cyber-defence applications. As such, there is a significant incentive for developing a quantitative metric for evaluating predicted successor representations in a cybersecurity context (i.e. ones that correspond to computer network topologies).

Considering the depth of insight provided by successor representations as compared to \tomnet's other two outputs further highlights this incentive. In contrast to next-step action probabilities and final target predictions, which constitute single-step predictions, successor representations capture the broader spatio-temporal aspects of an agent's policy in a single predictive map. In the context of cyber-defence, this could represent, for instance, an attacker's predicted trajectory through a network, unlocking several unique opportunities including subgoal discovery and the subsequent allocation of defensive resources to threatened intermediate services. The performance and effectiveness of such a capability would clearly demand rigorous and principled assessment.

We identify several requirements such a metric would have to fulfil to be of use to network administrators. It must, first of all, be interpretable. This rules out unbounded metrics that lack a fixed reference scale (e.g. the Mean-Squared Error), as human operators may not always possess the context or domain-expertise required to interpret their meaning, especially when comparing values for networks of different scales. Second, the metric must be able to handle sparse inputs. This is because cyber attackers can follow relatively narrow paths towards targets in the network, resulting in limited network exploration and, consequently, (near-)zero expected occupancies that can disproportionately affect the outputs of zero-sensitive metrics (e.g. the Kullback-Leibler Divergence, and, by extension, the Jensen-Shannon Divergence). Finally, given that computer networks, and therefore any corresponding successor representations, represent geometric spaces in which nodes adhere to specific topological configurations, having a metric that explicitly accounts for this is imperative. For instance, a predicted path that lies closer to a target path in terms of proximity in network space than another prediction is intuitively more useful, even if it is equally erroneous in terms of misplaced probability mass.


\medskip

\noindent In summary, what is needed to adapt the \mtom framework to the cybersecurity domain is:

\begin{enumerate}
    \item A flexible \tomnet archtecture that can maintain robust predictions over high-dimensional networks and feature spaces;
    \item An interpretable, stable metric for quantitatively evaluating predicted successor representations that respects the spatial relationships between their elements.
\end{enumerate}

%% file: sections/contributions.tex
\section*{Contributions Summary}

In this paper, we present a novel graph neural network (GNN)-based \tomnet architecture along with a specialised metric for measuring similarity between graphical successor representations.
Together these contributions address the above limitations to bring \mtom a step closer to being applicable to the domain of cyber-defence.

The literature on using machine learning for autonomous cyber-defence has identified a graph representation as a natural fit for the domain~\cite{palmer2023deep}. 
Therefore, we propose Graph-In, Graph-Out ToMnet (\gigo) for cyber-defence, for which both feature extractors and output layers are implemented using GNNs. 
\gigo provides a flexible formulation where the output dimensionality matches that of the presented graph-based observation, allowing it to parse inputs of variable dimensions. We train \gigo to reason over the likely behaviours and goals of cyber-attacking agents within a reputable, abstract cyber-defence environment \cite{YAWNING}, and show how these can be accurately predicted across a range of network topologies.

%
%

For our main contribution, we present a novel extension of the Wasserstein Distance (WD) \cite{rubner1998metric} for measuring similarity between graph-based probability distributions. The baseline Wasserstein Distance computes the minimum `cost' of translating one distribution into another by optimally reallocating probability mass according to a specified cost matrix $D$. Therefore, by setting $D$ to the matrix of pairwise shortest path lengths between nodes in the input graph, a value is generated representing the minimum amount of work required to equate input distributions by moving probability mass around the network. This value is upper-bounded by the network's diameter. Therefore, dividing it by the network's diameter yields a network-agnostic, unit-bounded metric representing a fraction of the worst-case scenario. We call this metric the \emph{\mtrclong} (\mtrc), and use it to evaluate predicted successor representations throughout our experiments.

For additional customizability, we furnish the \mtrc with a weighting function
that linearly combines an arbitrary set of node feature vectors
with a corresponding set of user-specified parameters
(including a set of weighting coefficients
for each selected node feature and a floor parameter
for the min-max scaling function
that determines the overall influence of the weighting on the final score). This yields a composite weights vector
that is subsequently used to rescale input distributions, thereby emphasizing areas of the network highlighted by the selected node features whilst preserving the metric's key theoretical properties.
We demonstrate how this method can be used to gain a strategically richer understanding of how prediction errors are distributed around the graph by analyzing discrepancies between the metric values produced by different weighting configurations.
%


%

%



%% file: sections/related_work.tex
\section*{Related Work} \label{sec:related_work}

GNNs are a class of neural networks
that can be applied to graphs--- data structures that systematically
model relationships between entities \cite{palmer2023deep}---making them
highly effective for applications where data points have explicit
relationships.
%
GNNs therefore represent a promising candidate for any \mtom problem that can
be represented graphically.
For example, Wang et al.~\cite{wang2021tom2c} used a GNN-based \tomnet to build socially intelligent agents.
Shu et al.~\cite{shu2021agent} evaluated if agents designed
to reason about other agents, including a GNN-based \tomnet
approach, can learn or hold the core psychology principles
that drive human reasoning.
GNN-based \tom approaches have also been applied to learn
the relationships between players within multi-player
games such as Press Diplomacy\cite{jeon2022inferring},
 and emergent adversarial communication \cite{piazza2023theory}.

As mentioned above, \tom has also previously been applied to cyber-security. 
%
Cheng et al.~\cite{9013291} introduced a \tom-based stochastic
game-theoretic approach to reason about the beliefs and behaviors of attackers by using a Bayesian attack graph to model multi-step attack scenarios.
%
%
Malloy and Gonzalez~\cite{10190658} present a novel model of human
decision-making inspired by the cognitive faculties of instance-based
learning theory, \tom, and transfer of learning, finding from experimentation in a simple Stackelberg
Security Game~\cite{von2010market} that theory of mind can improve
transfer of learning in cognitive models.
%
%
%
However, we have found no papers reviewing the general utility of \tomnet in the domain of cyber-defence. As such, to our knowledge, our broad focus on the
development of network-agnostic, scalable GNN-based \tomnet formulations is unique.

Although the application of our evaluation metric is specialized, there do exist promising candidates off-the-shelf.
%
In statistics, the Wasserstein Distance (WD), also known as the Earth Mover's Distance, is a measure used primarily for comparing probability distributions represented over a metric space~\cite{rubner1998metric}. It quantifies the minimal cost required to transform one distribution into another by considering the work needed to relocate the distribution mass in the most efficient manner possible.

Owing to its flexibility (especially with respect to being able to define a custom metric space), The WD has been extensively applied with various extensions to adapt it to different problem domains.
For instance, Wang et al.~\cite{wang2021normalized} introduced a normalized variant tailored for tiny object detection, which models bounding boxes as Gaussian distributions to improve the robustness of detection performance against minor localization errors. The normalization is key, as it converts a boundless distance metric to a standardised similarity measure (i.e. between 0 and 1, like IoU).

%
%
For graphical applications, Kim et al.~\cite{kim2004new} pioneered a two-step ARG matching algorithm that improved the robustness and performance of graph matching through nested WD computations, while Noels et al.~\cite{noels2022earth} crafted an WD-based graph distance metric tailored for analyzing financial statements, enhancing tools for fraud detection and company benchmarking.
Other closely related existing metrics in this domain include the Geometric Graph Distance (GGD)~\cite{cheong2009measuring}, which quantifies the minimum cost required to transform one geometric graph into another by modifying node positions and edge connections within a Euclidean space, and the Graph Mover's Distance (GMD) ~\cite{majhi2023graph}, which
offers a more computationally tractable extension of the GGD, simplifying the transformation process using the EMD and making it more feasible for large-scale applications.

While both GGD and GMD consider graph topology, their focus is primarily on geometric and structural transformations. Our method, the \mtrclong (\mtrc), distinguishes itself by focusing specifically on how information or resources flow through a network, directly incorporating paths and distances into the WD computation and subsequent normalization in a way that, to the best of our knowledge, has not been explored previously. Furthermore, in contrast to some of the more domain-specific WD extensions mentioned, the \mtrc offers a versatile metric that can be applied to any problem involving the comparison of graph-based probability distributions, and enriched with $\Wfn$ wherever one wishes to explore the relationship between node features and prediction errors.

In summary, whilst primarily geared at developing the field of cyber-defence, our methodology includes novel contributions to both the suite of existing \tomnet architectures as well as the network analysis toolbox that can be utilized beyond this domain.

%


%% file: sections/preliminaries.tex
\section*{Preliminaries} \label{sec:problem_forumulation}

\subsection*{The Hot-Desking User Problem} \label{subsec:hotdesk}
In this report we evaluate the ability of \tomnet architectures to characterize agents situated within cyber-defence games.
Our focus is on scenarios featuring two agents, a \red cyber-attacking agent and a \blue cyber-defence agent. 
For our specific use-case, we consider a scenario that we term the \emph{\hdu}. The problem consists of a computer network $\mathcal{G}$ with a static topology (\ie the set of nodes $v \in \mathcal{V}$ and the edges $e \in \mathcal{E}$ connecting them remains the same for a given topology). In each episode we have changes with respect to:

\begin{enumerate}[label=(\roman*)]
    \item the vulnerabilities of each node $v$ (\eg due to different services running);
    \item the location of the \hvns, which are determined based on the location of a set of users $u \in \mathcal{U}$.
\end{enumerate}


We assume that the number of users $\rvert \mathcal{U} \rvert$ remains consistent across episodes. However, due to hot-desking, the locations of the users change after each episode. Therefore, in each episode we have a different set of tuples $(u_i, v_j)$, where $i$ represents each user and $j$ represents each \hvn. In our scenarios, a \red cyber-attacking agent is attempting to evade a \blue cyber-defence agent in order to reach the \hvns, over which \red has a preference. This preference could be due to, say, a certain user working on a classified topic of interest to the \red agent. Based on the log-files from previous attacks---the past trajectories $\tauobs$---our \tomnet architectures are tasked with predicting:

\begin{enumerate}[label=(\roman*)]
    \item the exact node that \red is targeting:  $\TargetNode$;
    \item the attack trajectory followed by \red on the path towards $\TargetNode$ (or, the \emph{successor representation}): $\SR$ \cite{dayan1993improving}.
\end{enumerate}

\begin{figure}[h]
    \centering
    \includegraphics[width=\columnwidth]{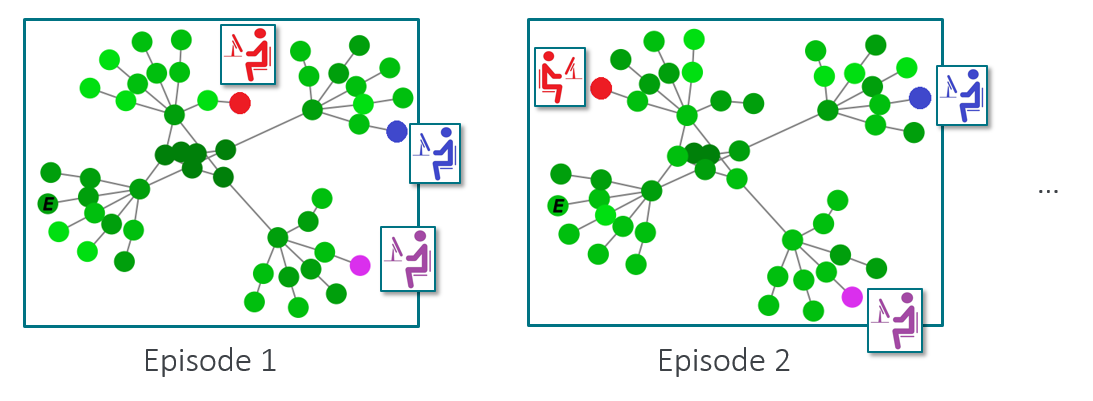}
    \caption{An illustration of the hot-desking user problem for cyber-defence. The node labelled `$E$' denotes the entry node for the \red cyber-attacking agent. Blue, red, and pink nodes represent current user locations. The color weighting of the remaining nodes represents their vulnerability to attacks, with darker shades representing higher vulnerability scores.
    In each episode the users may relocate to a different desk.
    Therefore, \tom solutions are required that can generalize across settings
    with respect to different user locations in each episode.}
    \label{fig:hotDeskingUserProblem}
\end{figure}

\noindent An illustration of the \hdu is provided in \autoref{fig:hotDeskingUserProblem}.

\subsection*{Machine Theory of Mind} \label{sec:background:mToM}

The \hdu can be thought of as a \emph{partially observable Markov game} (POMG). Unlike in (fully observable) Markov Games, also known as stochastic games \cite{shapley1953stochastic}, the full state of the environment is hidden from all players within a POMG. This closely reflects the conditions of real-world cyber-defence scenarios, wherein cyber-attacking agents will typically be unable to observe the complete layout of the network and all the services running on each node. 
Similarly, cyber-defence agents will rarely be able to observe the full state of the network given the computational overhead that would be required to relay all of this information at every time-step.

Formally, our task for \tomnet is to make predictions over a family of POMGs:
\begin{equation}
    \M = \bigcup_j \M_j.
\end{equation}
Each $\M_j$ represents a POMG and is
%
defined as a tuple $(\mathcal{N}, \states, \actions, \transition, \rewards, \gamma, \observations, \observationf)$,
where:
\begin{itemize}

    \item $\mathcal{N}$ is a set of agents.

    \item $\states$ is a finite state space.

    \item $\actions$ is a joint action space $(\actions_1 \times ... \times \actions_n)$ for each state $\state \in \states$, with $\actions_i$ being the number of actions available to player $i$ within a state $\state$. 
    \item $\transition$ is a state transition function: $\states_t \times \actions_1 \times ... \times \actions_n \times \mathcal{\states}_{t+1} \rightarrow [0,1]$, returning the probability of transitioning from a state $\state_t$ to $\state_{t+1}$ given an action profile $\action_1 \times ... \times \action_n$.
    Here, each action $\action_i$ belongs to the set of actions available to agent $i$: $\action_i \in \actions_i$.
    %

    \item $\observations$ is a set of 
observations.

    \item $\observationf$ is an observation probability function defined as
    $\observationf : \states \times \actions_1 \times ... \times \actions_n \times \observations \rightarrow [0,1]$, where,
    for agent $i$, a distribution over observations $\observation$ that may occur in state $\state$ is returned, given an action profile $\action_1 \times ... \times \action_n$.

    \item $\mathcal{\rewards}$ is a reward function: $\mathcal{\rewards} : \mathcal{\states}_t \times \mathcal{\actions}_1 \times ... \times \mathcal{\actions}_n \times \mathcal{\states}_{t+1} \rightarrow \rats$, that returns a reward $r$. 

    \item $\gamma$ is a discount rate defining the agent's preference over immediate and future rewards within the decision-making process. Larger values for $\gamma$ define a preference for long-term rewards.

\end{itemize}
For the environments considered in this paper we also
allow \emph{terminal states} at which the game ends.

The \tomnet architectures in this report are tasked with making
predictions over a range of network topologies with different node settings.
We also assume we have a family of agents:
\begin{equation}
    \agents = \bigcup_i \agents_i,
\end{equation}
with each $\agents_i$ representing a specific agent.
Each agent is implemented by a policy $\pi_i$,
defining the agent's behaviour.
We shall use the same problem formulation as Rabinowitz et al. \cite{rabinowitz2018machine},
and associate the reward functions, discount factors,
and conditional observation functions with the agents $\agents$ rather
than the POMGs $\M$.
Therefore, for each POMG $\M_j$ we have a tuple
$(\states, \actions, \transition)$, and for each agent $\agents_i$ we have
$(\observations, \observationf, \rewards, \gamma, \pi)$.

%
%
%
For our \tomnet approaches we assume access to (potentially partial and/or noisy)
observations of the agents, via a state-observation function
$\stateobsfun(\cdot) : \states \rightarrow \Omega^{\obs}$
and an action-observation function
$\actionobsfun(\cdot) : \actions \rightarrow \action^{\obs}$.
For each agent $\agents_i$ in a POMG $\M_j$, we assume our observer
can see a set of \emph{trajectories}, defined as sets of state-action pairs: $\tauobs_{i,j} = \{(\stateobs, \actionobs)\}^{T}_{t=0}$,
where $\actionobs = \actionobsfun(a_t)$ and $\stateobs = \stateobsfun(s_t)$.
%
%
%
%
Critically, setting $\stateobsfun(\cdot)$
and $\actionobsfun(\cdot)$ to the identity function grants the observer
full observability of the POMG state along with all overt actions taken
by the agents while keeping their parameters,
reward functions, policies, and identifiers concealed.
This formulation closely approximates the depth of knowledge
that would be available to a human observer in a real-world
scenario.

Our approaches are based on the original \tomnet \cite{rabinowitz2018machine} (\autoref{fig:ToM}).
To recap, \tomnet consists of three modules, a
\emph{character net},
\emph{mental net}, and
a \emph{prediction net}.
The character net $f_{\theta}$ yields a character embedding
$\characteri$ upon parsing $N_{past}$ past trajectories $\tauobs$
for an agent $\agents_i$, $\{\tauobs_{i,j}\}^{N_{past}}_{j=1}$.
As in the original paper, we shall process the
trajectories independently using a
recurrent neural network,
implemented with a
Long Short-Term Memory (LSTM) network \cite{hochreiter1997long}
$f_\theta$, and sum the outputs:
\begin{equation} \label{eq:characteri}
\characteri = \sum^{N_{past}}_{j=1}f_\theta \Bigbrackets{\tauobs_{ij}}.
\end{equation}
The mental (state) net $g_{\vartheta}$ meanwhile processes the current
trajectory up to time-step $t-1$ along with $\characteri$:
\begin{equation}
\mentali = g_{\vartheta}\Bigbrackets{[\tauobs_{ij}]_{0:t-1}, \characteri}.
\end{equation}
The prediction net subsequently uses both embeddings and the current
state $\state$ to predict the agent's behaviour, including: items
that the agent will consume $\hat{c}$ (\eg a \hvt node that
a \red cyber-attacking agent wants to gain access to), and; the successor
representations $\SR$ \cite{dayan1993improving} (i.e. the \red cyber-attacking agent's attack trajectory towards $\hat{c}$).

\subsection*{Graph-Based \yt} \label{sec:problem_forumulation:gyt}

We simulate the \hdu in an adapted version of the \yt cyber-defence environment. This is an open-source framework originally built to train cyber-defence agents to defend arbitrary network topologies \cite{YAWNING}. Each machine in the network has parameters that affect the extent to which they can be impacted by \blue and \red agent behaviour. These include vulnerability scores that determinine how easy it is for a node to be compromised. Our extension includes graph observations for both graph-based ToMnet architectures and graph-based Blue cyber-defence agents.
Critically, these observations will contain different combinations of node features depending on the observability of the agent being trained. For example, a \tomnet architecture with full observability will receive observations containing relevant labels in the node features (such as \hvns), whereas these will be absent for cyber agents with partial observability.

\subsubsection*{Networks} \label{sec:problem_forumulation:gyt:networks}

\begin{figure}[ht]
    \centering
    \subfloat[\treeThirty]{
    \label{fig:treeThirty}
    \includegraphics[width=0.25\columnwidth]{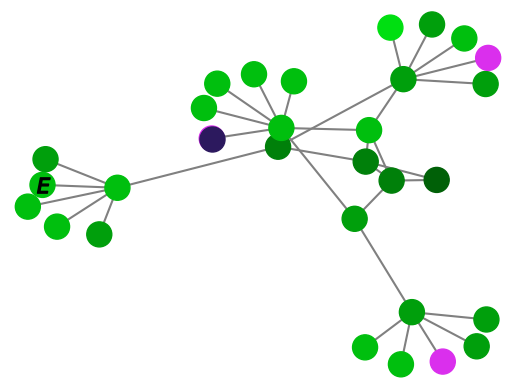}}
    \subfloat[\treeForty]{
    \label{fig:treeForty}
    \includegraphics[width=0.25\columnwidth]{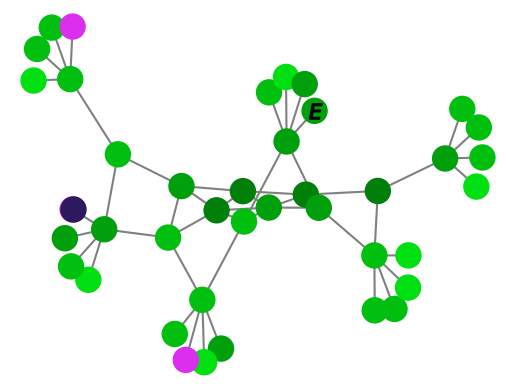}}
    \subfloat[\treeFifty]{
    \label{fig:treeFifty}
    \includegraphics[width=0.25\columnwidth]{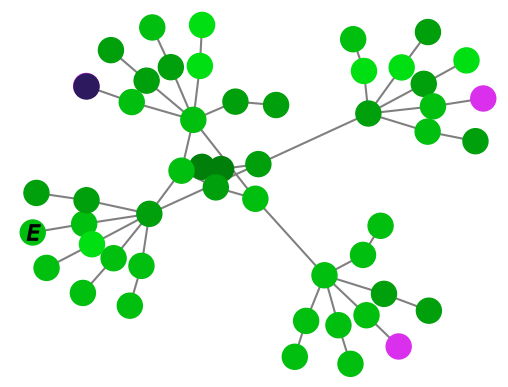}}

    \subfloat[\treeSeventy]{
    \label{fig:treeSeventy}
    \includegraphics[width=0.25\columnwidth]{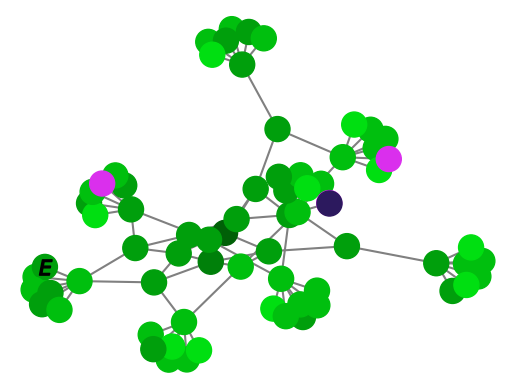}}
    \hspace{10mm}
    \subfloat[\treeNinety]{
    \label{fig:treeNinety}
    \includegraphics[width=0.25\columnwidth]{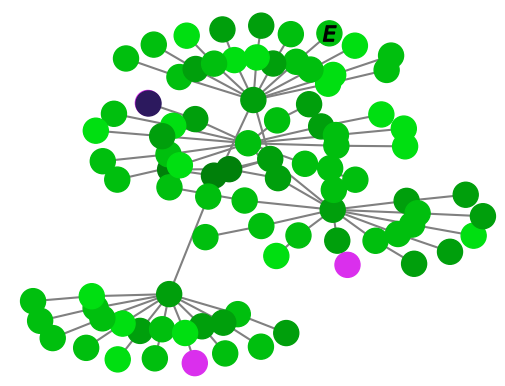}}

    \caption{Example visualizations of the five custom \yt \tree topologies used in our experiments. For each topology, the node labelled `$E$' denotes the entry node for the \red agent. Pink nodes represent \hvns, with the dark blue node being the one ultimately targeted by a hypothetical \red agent.
%
}
    \label{fig:NetworkTopologies}
\end{figure}

Custom network configurations are required to systematically evaluate \gigo across different network sizes and topologies. Therefore,
we have added a new network generator, \tree,
which resembles a tree network with layers for the core, edge,
aggregation, access, and subnet nodes and can be easily visualized (\autoref{fig:treeThirty} -- \ref{fig:treeNinety}).

%

%

\subsubsection*{Agents}

To enable an extensive evaluation, 
we have additionally implemented a range of rule-based \blue
cyber-defence and \red cyber-attacking agents to our
graph-based \yt framework.
While our agent repertoire is broad (a comprehensive overview is provided in 
\autoref{app:agents}), 
our evaluations suggest that only two were useful in generating interesting behaviour for our experiments:

\begin{itemize}
    \item \blueMSND:
    A deterministic agent that removes the infection from a compromised node (via the \msn action) if it is within three hops of any \hvn,
    otherwise scans the network for hidden compromised nodes (via the \scan action).
    The strategy here is to protect only the core of the network. \blue is content with sacrificing the edge nodes to do so.

    \item \redHVTPreferenceSP: Selects a particular \hvn to target at the start of the episode and takes the shortest path towards it.
    The selected \hvn is based on a \hvn preference vector sampled from a Dirichlet distribution $\pi \sim Dir(\alpha)$ as well as
    the shortest distance between \hvns and entry nodes.
    Therefore, following \cite{rabinowitz2018machine}, we can define an agent \emph{species} as corresponding to a particular value of $\alpha$, with its members being the various parameterizations of $\pi$ sampled from $Dir(\alpha)$.
    This attacker has prior knowledge of the network structure, for example, due to insider information,
    meaning it can calculate path lengths and take the most direct path to its chosen high-value node.
    \end{itemize}

The motivation for using rules-based agents over incentive-based learning approaches lies in
their interpretability. For example, we can define a preference over
high-value targets and actions, as well as reducing the amount
of time required for gathering training and evaluation data
for our experiments.
%
%
However, we note that, in principle, \tom approaches are agnostic with respect to how
the \blue and \red agent policies are obtained, and can
therefore also be applied to learnt policies.

%% file: sections/methods.tex
\section*{Methods}  

As previously mentioned, one of the challenges when applying \tomnet to
cyber-defence scenarios is that that the number of
nodes in the network is not necessarily known in advance.
This has implications for both the input and output
layers of \tomnet.
For our current problem domain, neural network layers
are required that can handle variable sized inputs~\cite{palmer2023deep}.
There are now a number of solutions to this
problem.
For example, using transformers \cite{vaswani2017attention}
or applying one dimensional convolutions with global max
pooling.
However, we consider that a natural format for representing
a cyber-defence scenario on a computer network is to use a graph-based
representation.
Therefore, 
we have 
implemented a graph-based observation wrapper for the YAWNING-TITAN
cyber-defence environment~\cite{YAWNING}.

In order to process the graph-based state-observations,
we introduce two graph-based \tomnet architectures for
making predictions for cyber-defence scenarios:
i.) a \emph{Graph-in, Graph-Out} \tomnet (\gigo), where
both input and output layers are implemented using graph neural networks, and;
ii.) a \emph{Graph-in, Dense-Out} \tomnet (\gido), where inputs are processed
by graph neural network layers and outputs are generated by subsequent dense
neural network layers, more closely approximating the original \tomnet formulation to benchmark our evaluations.

Following evidence from the literature that graph attention layers \cite{GAT} are less susceptible to the `over-squashing' of node features than other popular candidates \cite{alon2020bottleneck}, these are used for node feature extraction end-to-end in both models.

For both \gido and \gigo, the character network processes
sequences of graph-based state-observations $\tauobs_{ij}$ called \emph{trajectories}.
From the trajectories, each graph is fed through two GATv2 layers
combined with dropout layers ($p=0.5$).
The outputs from each layer are then subjected to
both \emph{global max} and \emph{global average} pooling, with the
outputs subsequently being concatenated.
The resulting features for each graph are then sequentially fed into
an LSTM that generates a character embedding $\characteri$ for each
individual trajectory $i$.
The outputs from the LSTM are then summed as described in
\autoref{eq:characteri},
resulting in the character embeddings.

\begin{figure}[h]
    \centering
    \includegraphics[width=0.95\columnwidth]{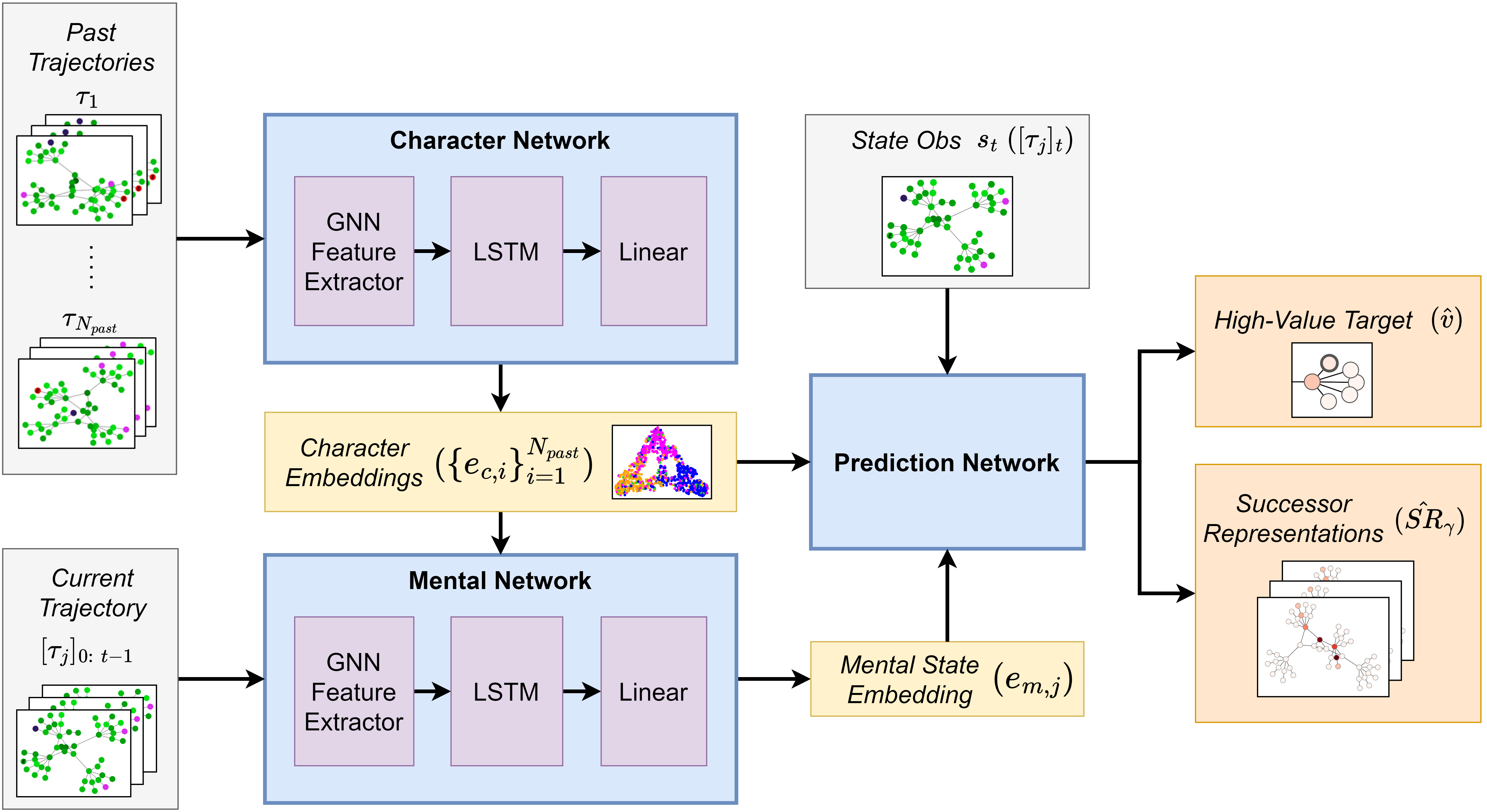}
    \caption{Shared architectural components of \gigo and \gido.}
    \label{fig:shared_arch}
\end{figure}

GATv2 layers are also used to extract features for the
mental embedding $\mental$ and for extracting features that
are fed into the prediction network.
While our \gido and \gigo architectures share the same
feature extraction for the character and mental network components (\autoref{fig:shared_arch}),
 they differ with respect to the implementation of the prediction network
and output layers.
Below we describe these differences in more detail.

\subsection*{Graph-In, Dense-Out ToMnet} \label{sec:methods:GIDO}

Similar to the original \tomnet, for \gido, the character
and mental embeddings, $\character$ and $\mental$, along with the current
state observation $\stateobs$, are fed into a prediction network.
The prediction network first uses GATv2 layers
to extract features from the $\stateobs$, before concatenating
the resulting embedding with $\character$ and $\mental$.
%
%
The concatenated embeddings are subsequently fed through a
number of dense layers before being passed to the individual output
layers: i.) a prediction over the consumable $\hat{c}$, representing
a prediction over which \hvt the \red cyber-attacking
agent wants to reach, and; ii.) the successor representation $\SR$.
%
%
%

As for the original \tomnet \cite{rabinowitz2018machine}, past and current
trajectories will be sampled for each agent $\agents_i$ to obtain predictions.
Each prediction provides a contribution to the loss, where the average of
the respective losses across each of the agents in the minibatch is used
to give an equal weighting to each loss component.

For the \hvt losses we use the negative log-likelihood
loss for each \hvn $n$:
\begin{align}
\Lhvti = \sum_n - \log p_{c_n} \bigbrackets{c_n \rvert \stateobs, \characteri, \mentali}.
\end{align}

\begin{figure}[H]
    \centering
    \includegraphics[width=0.75\columnwidth]{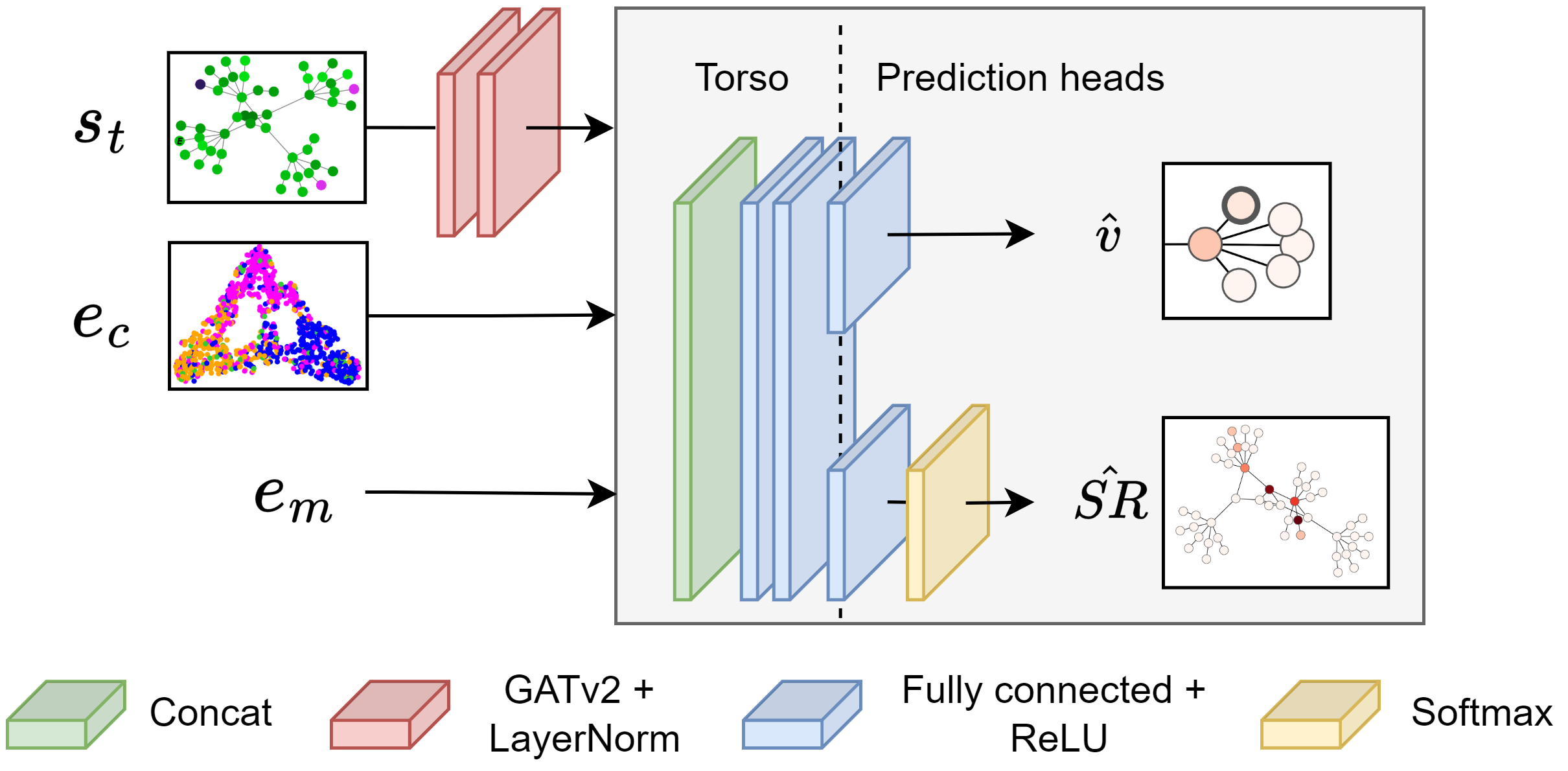}
    \caption{The Graph-In Dense-Out (\gido) prediction network architecture. Output layers are implemented using dense neural network layers, necessitating a fixed number of output nodes.}
    \label{fig:gido}
\end{figure}

Finally, we apply a soft label cross entropy loss $\Lsr$ to optimise the network with respect
to predicting the successor representations:
\begin{equation} \label{eq:sri_loss}
\Lsri = \sum_{\gamma} \sum_{s} - SR_\gamma(s) \log \SR_\gamma(s),
\end{equation}
where $s$ is a state-observation $\stateobs$, and for each discount factor $\gamma$, we
obtain the ground truth $SR_\gamma(\stateobs)$ through an empirical normalised discounted
rollout, from the time-step $t$ where the state-observation was gathered, onwards.
The combined loss is:
\begin{equation} \label{eq:total_loss}
\mathcal{L}_{total} = \Lhvt + \Lsr.
\end{equation}
The disadvantage of the above approach is that we fix the max number
of nodes per output layer through examining the maximum number of nodes present
within the dataset.
\gido is depicted in \autoref{fig:gido}.

\subsection*{Graph-In, Graph-Out ToMnet} \label{sec:methods:GIGO}

In contrast to \gido, for \gigo we concatenate the character and mental
network embeddings directly with each node's features within the current
state-observation.
These are then directly processed by GNN layers before each
of the prediction tasks.
As mentioned above, for \gigo, the \hvt and successor representation predictions are generated using GATv2 layers.
The final GNN layer of each respective prediction branch represents the outputs,
\ie
a weighted binary cross entropy loss is applied to each node for optimizing high
value node predictions:
\begin{equation}
\Lhvti = \sum_n - w_n (y_n \log x_n + (1 - y_n) \log(1-x_n)).
\end{equation}
where $x$ represents the ground-truth with respect to whether or not a node
was consumed at the end of the episode, and $y$ is the prediction.
The positive weighting term $w_n$ is added to account for the positive / negative
node imbalance \emph{within} an individual sample, i.e., only
one out of $n$ nodes will (potentially) be consumed at the end of a given
episode.

For the successor representation loss $\Lsr$ we stick with the soft label cross
entropy loss from Rabinowitz et al. \cite{rabinowitz2018machine}.
%
For batches where the number of nodes on each graph is inconsistent we zero-pad
while computing the loss.
We note this formulation only requires padding when computing the loss, meaning at
inference time there are no superfluous nodes (in contrast to when specifying the
max number of nodes for \gido).

\begin{figure}[h]
    \centering
    \includegraphics[width=0.75\columnwidth]{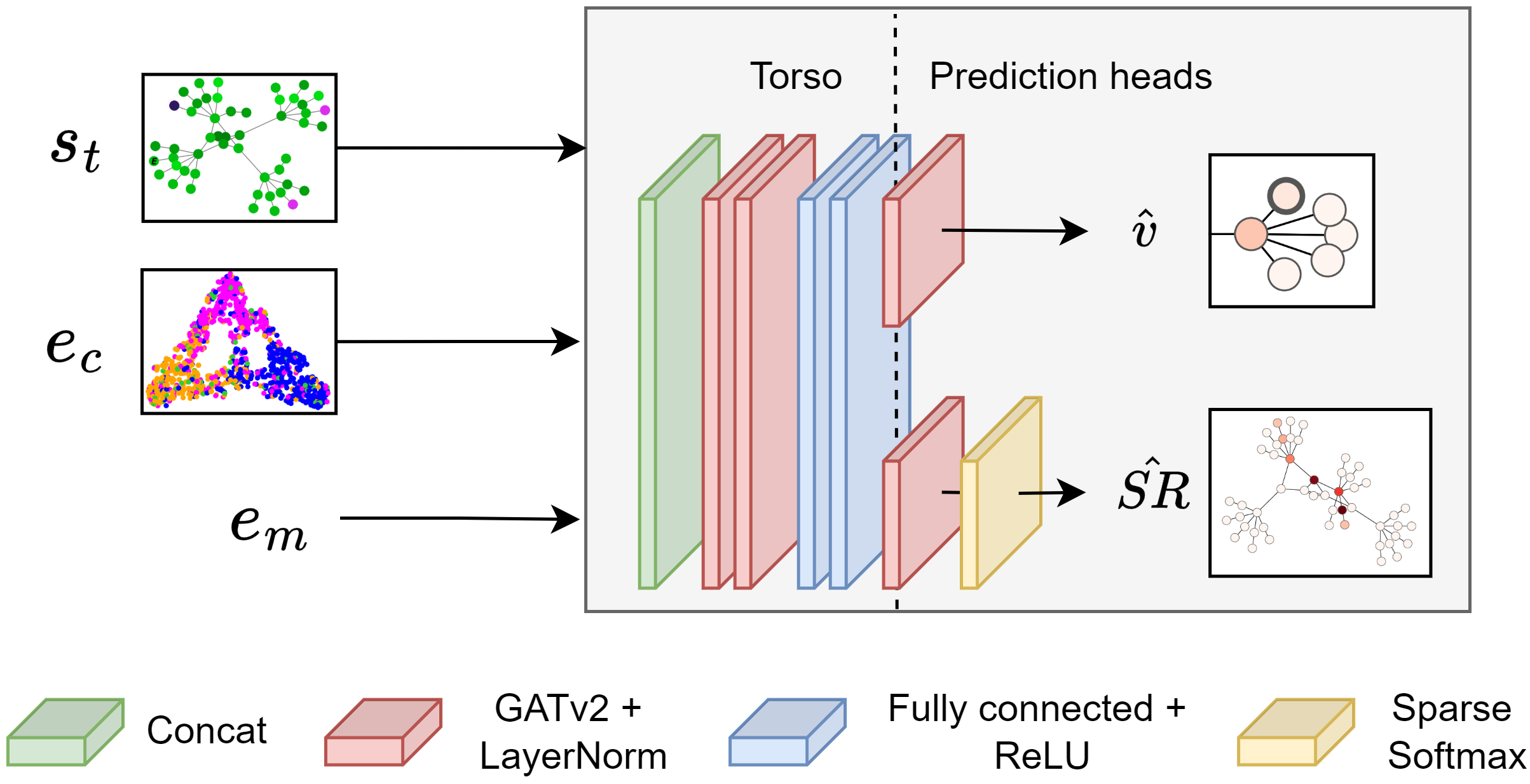}
    \caption{The Graph-In Graph-Out (\gigo) prediction network architecture. Both feature extractors and output layers are implemented using graph neural network layers, providing a flexible formulation that can parse inputs and generate outputs of variable dimensions.}
    \label{fig:gigo}
\end{figure}

The approach outlined above provides a flexible formulation regarding the number of nodes in the
network, with the output topology reflecting that of the current state-observation fed into the
prediction net.
\gigo is depicted in \autoref{fig:gigo}.

\subsection*{The \mtrclong} \label{sec:sr_metric:methods}

Of the \tomnet outputs, one that is particularly salient for the depth of insights on offer to network administrators is the \textit{successor representation}.
In the field of reinforcement learning, these are formally defined as matrices representing the expected discounted future state occupancy following a given policy from a given state \cite{dayan1993improving}.
In the case of our \hdu, these represent predictions regarding the likely attack trajectory that Red will take against the current Blue cyber-defence agent.
Predicted successor representations are often evaluated using generic
metrics that assume input data points are independent and identically distributed (i.i.d.).
However, in real-world cyber-defence scenarios, the nodes of a network will often have individual features granting them varying degrees of vulnerability, utility, and overall strategic importance \cite{doshi2017towards}, and will adhere to a certain topological configuration. As such, a metric for evaluating predicted successor representations that captures these contextual nuances is desirable.

We achieve this with the \mtrclong (\mtrc); a unit-bounded extension of the Wasserstein Distance (WD) that includes a graph-theoretic normalization factor that represents the result as a fraction of the worst-case predictive scenario for the network in question.

The metric can be calculated as follows:
\begin{equation}
    \text{\mtrc}(P,\: Q,\: D) = \frac{1}{\text{max}(D)}\inf_{\mu \in M(P, Q)} \left[ \int_{X \times X} d(i, j) \; d\mu(i, j) \right],
\end{equation}
where:
\begin{itemize}
    \item $P$ and $Q$ represent probability distributions in which individual elements correspond to nodes in a graph $\graph$.
    \item $D$ represents the matrix of pairwise shortest path lengths between nodes in $\graph$. The diameter of $\graph$ is therefore $\text{max}(D)$.
    \item $\inf_{\mu \in M(P, Q)}$ denotes the infimum (i.e. the greatest lower bound) over all possible transport plans $\mu$. $M(P, Q)$ represents the set of all joint distributions whose marginals are $P$ and $Q$.
    \item $\int_{X \times X} d(i, j), d\mu(i, j)$ calculates the total cost of transporting mass from distribution $P$ to distribution $Q$. The function $d(i, j)$ is the ground metric on the graphical space $X$, which here is the shortest path length between nodes $i$ and $j$ in $G$. Therefore, $d(i, j) = D[i, j]$. $d\mu(i, j)$ denotes the amount of probability mass transported from $i$ to $j$.
\end{itemize}

Like the Wasserstein Distance (WD), the \mtrc is symmetric and non-negative:
\begin{equation}
    \text{\mtrc}(P,\: Q,\: D) = 0 \Leftrightarrow P=Q.
\end{equation}
Unlike the WD, however, the \mtrc has an upper bound of 1:
\begin{equation}
    \text{\mtrc}(P,\: Q,\: D) = 1 \Leftrightarrow \text{WD}(P,\: Q,\: D) = \text{max}(D).
\end{equation}

\begin{figure}[h]
    \centering
    \includegraphics[width=.8\columnwidth]{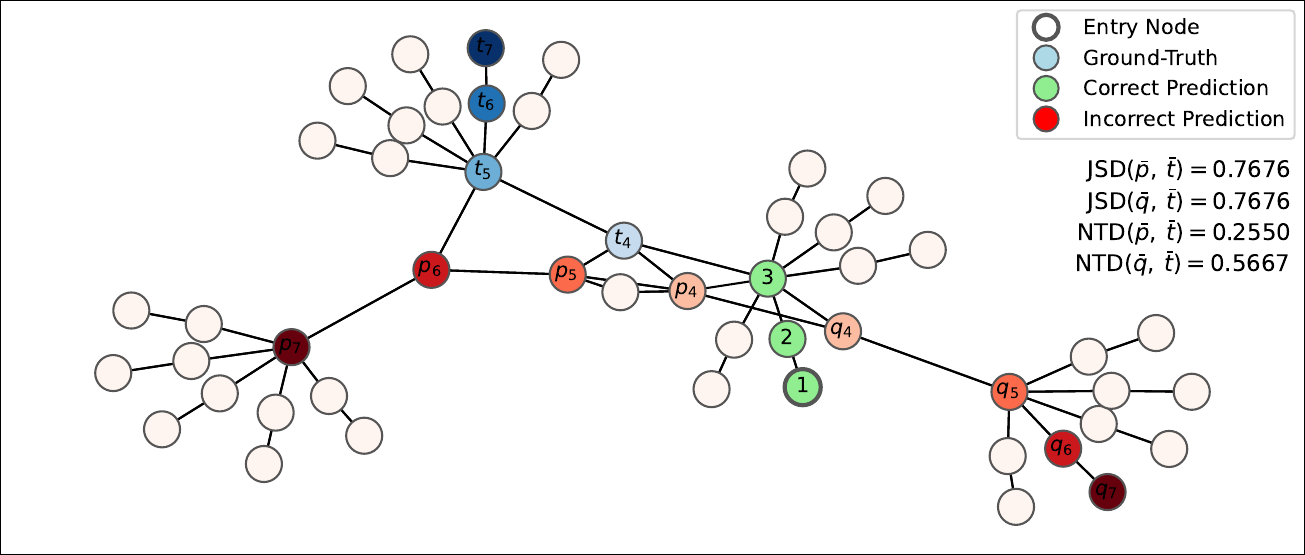}
    \caption{
        Hypothetical example demonstrating the importance of a topology-aware  metric for evaluating graph-based successor representations. Shown is a 50-node network with three distinct attack paths overlaid: a ground-truth $\bar{t}$, and two predictions $\bar{p}$, $\bar{q}$. For demonstrative purposes, each path is the same length and each node along the paths is assigned the same probability score, with only their locations varying i.e. $\forall i, \: t_i = p_i = q_i$. Therefore, depicted is a scenario where $\bar{p}$, $\bar{q}$, and $\bar{t}$ represent different graphical translations of the same probability distribution. Green and white nodes represent locations where probability masses for all three distributions are identical, with green denoting shared non-zero values and white shared zero values. Erroneous predictions are displayed in red, and the ground-truth in blue. Darker shades indicate higher probability scores. \mtrc scores for each predicted path are displayed on the right, along with Jensen-Shannon distance (JSD) scores for a non-parametric comparison. All three paths coincide for for the first 3 nodes before diverging into different regions of the network. However, the erroneous segment of $\bar{p}$ is considerably `closer' to $\bar{t}$ than that of $\bar{q}$ in terms of node proximity in the graph space. In contrast to the JSD, which ignores this information, the \mtrc explicitly leverages the graph's structure to yield $\text{\mtrc}(\bar{p},\: \bar{t}) < \text{\mtrc}(\bar{q},\: \bar{t})$, as desired.
    }
    \label{fig:geometry_metric}
\end{figure}

This makes the metric network-agnostic and more interpretable to a broad audience, including non-technical stakeholders.
An illustrative example of why incorporating topological information into the \mtrc computation can be seen in \autoref{fig:geometry_metric}.

For added contextual awareness, we weight the \mtrclong by selecting a set of node features from a network based on their relevance to the strategic importance of node attacks, then linearly combining them with a corresponding set of user-specified coefficients using a customizable weighting function that can be applied to arbitrary metrics. We denote the end-to-end metric $\srsq$.

\begin{figure}[H]
    \centering
    \includegraphics[width=.98\columnwidth]{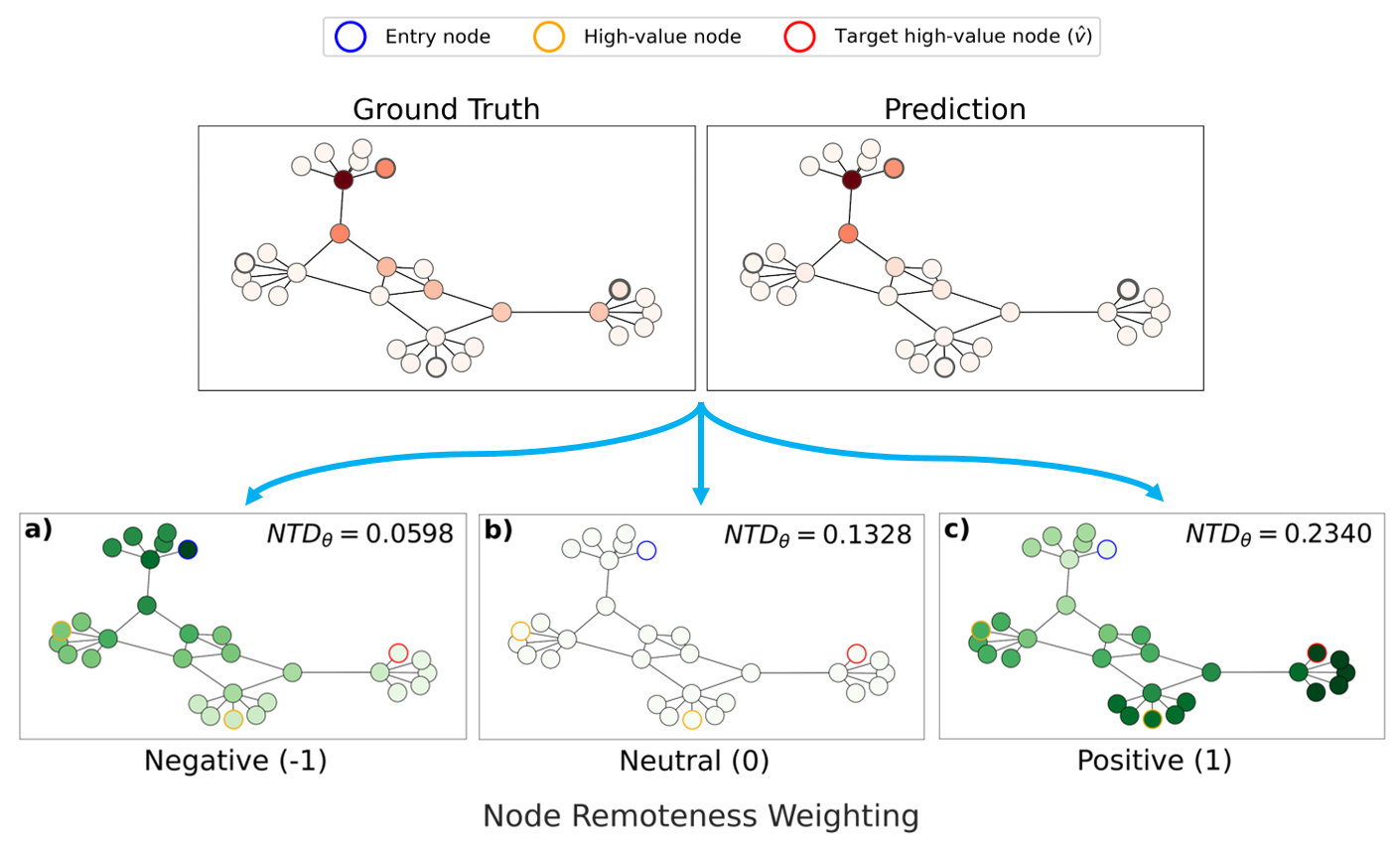}
    \caption{
        An illustrative example showing the influence of our weighting function $\Wfn$ on output $\srsq$ scores under different node feature weighting parameterizations. The top figures display a ground truth and predicted attack path ($\SR_{true}$ and  $\SR_{pred}$) for \treeThirty. Darker shades of red represent higher likelihoods of a node coming under attack. The three plots beneath (\textbf{a}-\textbf{c}) show the composite node weights $w_i \in \w$ and corresponding $\srsq\left(\SR_{true},\: \SR_{pred}\right)$ scores under different weighting parameterizations $\theta$. Higher weights are observable via darker shades of green. For this example, we define a single \emph{node remoteness} feature as the shortest path length from each node to the entry node. Varying the coefficient of this feature in $\Wfn$ therefore emphasizes different parts of the network based on proximity to the entry node. Subfigure \textbf{a)} shows a negative weighting, which emphasizes predictions closer to the entry node; \textbf{b)} shows the unweighted \mtrc, and; \textbf{c)} shows a positive weighting that emphasizes predictions made further away from the entry node. The significant $\srsq$ increase observed when inverting the coefficient from negative to positive reveals that $\SR_{pred}$ diverges further from $\SR_{true}$ with distance from the entry node, as can be visually confirmed in the upper figures. Such tests over $\srsq$ scores averaged across larger batches of predictions yield more general insights into predictive idiosyncrasies.
        }
    \label{fig:sr:example_sr_w_comparison}
\end{figure}

\noindent Mathematically, for a network $\graph$ with $n$ nodes, we have:
\begin{itemize}
    \item A set of $m$ user-selected node feature vectors: $\feats = [\feat{0}, \ldots, \feat{m}] \in \mathbb{R}^{n \times m}$.
    \item A corresponding set of $m$ user-specified feature weights: $\featsw = [\featw{0}, \ldots, \featw{m}] \in [-1, 1]^m$.
\end{itemize}
Our weighting function $\Wfnfull$ linearly combines these and normalizes the result:
\begin{equation}
    \label{eq:sr_weighting}
    \mathcal{W}(\mathcal{X}, \mathcal{C}, f) = \minmax{\sum_{i=0}^{m} \featw{i}\minmax{\feat{i}}{f}}{f} = \bar{w},
\end{equation}
where:
\begin{itemize}
    \item $\feat{i} \in \feats$ is a node feature vector.
    \item $\featw{i} \in \featsw$ is a user-specified constant that determines the weight of $\feat{i}$ in the final metric ($-1 \leq \featw{i} \leq 1$).
    \item $\minmax{\cdot}{f}$ is the min-max scaling function with `floor' parameter $f$, with $f$ defining the minimum weight that can be assigned to an input value ($0 \leq f \leq 1$). This forces all outputs to lie in the interval $[f, 1]$, thereby guaranteeing a minimum level of influence on the output metric:
    \begin{equation}
        \label{eq:minmax}
        \minmax{\bar{x}}{f} = \frac{(\bar{x} - \text{min}(\bar{x}))(1-f)}{\text{max}(\bar{x}) - \text{min}({\bar{x}})} + f.
    \end{equation}



\end{itemize}
The computed weighting $\bar{w}$ is used to rescale the input probability distributions before passing them into the \mtrclong. Therefore, for two $\graph$-based probability distributions $P, Q \in \mathbb{R}^n$, we have:
\begin{equation}
    \srsq(P, Q) = \text{NTD}(P_{\theta}, Q_{\theta}),
\end{equation}
where $\bar{x}_{\theta} = \frac{\bar{w} \cdot \bar{x}}{\sum \bar{w} \cdot \bar{x}}$.
It follows that $\Wfn$ is theoretically decoupled from the metric it is applied to, and can therefore be applied to any metric that is amenable to input weighting.
An illustration of influence of our weighting function $\Wfn$ on output $\srsq$ scores under different node feature weighting parameterizations can be seen in \autoref{fig:sr:example_sr_w_comparison}.

%% file: sections/results.tex
\section*{Experiments} \label{sec:experiments}

In this section, we empirically evaluate \gigo's ablity to provide actionable insights into the characters, goals, and behaviours of various cyber-attacking agents in the \yt environment, using the \hdu as our task formulation.

First, we explore \gigo's ability to characterize the policies of unseen cyber-attacking agents. We next investigate how well \gigo can predict the services targeted by these agents as well at their attack routes towards them.

Benchmarked against our dense-layered \tomnet implementation (\hyperlink{sec:methods:GIDO}{\gido}), we find that, through observing past episodes of \blue and \red agents interacting with each other, \gigo can learn character embeddings that can effectively differentiate between \red agents' preferred outcomes, providing a means to characterize their policies. We also observe that GIGO-ToM can accurately predict \red's preference over \hvns as well as their attack trajectories towards them, occasionally predicting multiple attack paths in cases where the target node is uncertain.

\medskip

\noindent We now describe our experimental set-up before presenting these results in detail.

\medskip

\noindent \textbf{Games:}
For the following experiments, we consider games consisting of the \blueMSND cyber-defence agent against 1,000 possible parameterizations of the \redHVTPreferenceSP cyber-attacking agent. The policy of each \red agent $\agents_i$ is defined by a fixed vector $\pi_i$ specifying its preference over high-value targets. These are sampled from a single agent species: a Dirichlet distribution with concentration parameter $\alpha=0.01$, ensuring a diverse range of sparse policies. For generality, we build a mixed topology setting named \treeMixed for which each game is rolled out in one of the following network topologies:  \{\hyperlink{fig:treeThirty}{\treeThirty}, \hyperlink{fig:treeForty}{\treeForty}, \hyperlink{fig:treeFifty}{\treeFifty}, \hyperlink{fig:treeSeventy}{\treeSeventy}, \hyperlink{fig:treeNinety}{\treeNinety}\}. Each network has three high-value users/nodes.
The location of the \hvns (and therefore the
location of the \hvus) are randomly selected at
the start of each episode. 
One entry node is provided to the \red cyber-attacking agent.
Given that our focus is on our models' ability to predict
\hvn locations, we simplify the
task with respect to the entry node by keeping it consistent across episodes.
To ensure a dataset of distinctive
trajectories for our experiments, \hvns are always situated
on leaf nodes.

\medskip

\noindent \textbf{Data:} To build a \tomnet dataset $\mathcal{D}_{ToM}$, a set of \blue cyber-defence agents $\agents^{blue}$ and a set of \red cyber-attacking agents $\agents^{red}$ is defined, along with a set of network configurations $\M$ for them to compete within. The Cartesian product of these three sets defines a set of game configurations: $G = \agents^{blue} \times \agents^{red} \times \M$. For each game $g \in G$, $n_c$ current episodes are generated, for each of which a distinct set of $n_p$ further past episodes is generated so that the sets of past trajectories $\mathcal{P}_i$ for all samples $s_i \in \mathcal{D}_{ToM}$ are mutually exclusive: $\bigcap^{|\mathcal{D}_{ToM}|}_{i=1} \mathcal{P}_i = \emptyset$. This precludes data leakage between training and validation sets inflating performance metrics. Thus, $n_{total} = n_c + n_c n_p$ episodes are generated for each game configuration $g \in G$, resulting in a database of trajectories: $\bigcup_{i=1}^{|G|} \{\tau_{ij}\}_{j=1}^{n_{total}}$.
We fix $n_c=3$ and $n_p=8$ for all experiments, generating 27 total trajectories for each game configuration $g \in G$.
Therefore, using a single \blueMSND agent against 1,000 parameterizations of the \redHVTPreferenceSP agent in the \treeMixed topology setting, a set of 3,000 current episodes are generated along with a pool of 24,000 episodes from which past and current trajectories could be extracted for building data samples. Each data sample $s_i \in \mathcal{D}_{ToM}$ consists of $N_{past}$ past trajectories for the character network, a current trajectory up to a certain time-step $t$ for the mental network, and the current state observation at that step $\stateobs$ for the prediction network, along with any ground-truth variables required for evaluation (e.g. $\TargetNode_{true}$ and $\SR_{true}$).
%
%
Given that trajectories often comprise hundreds of time-steps under our settings, we sample state-observations from past trajectories at regular intervals to reduce wall-times.
Five state-observations were sampled from each past trajectory for the following experiments.
For each current trajectory parsed by the prediction network, we uniformly select state observations $\stateobs$ from the first two time-steps of the current episode.
We split our resulting dataset into train and validation sets according to the ratio of 75\% / 25\%.

\medskip

\noindent \textbf{Evaluation:} To evaluate GIGO-ToM, we systematically examine each of its modules' outputs over a hold-out test set of 200 unseen \red agents. Character networks are qualitatively evaluated through inspecting the embeddings $\characterfull$ they generate for each agent. We collect the embeddings obtained from our test samples and apply t-SNE \cite{van2008visualizing} in order to obtain 2D visualizations of the clusters that emerge. Prediction networks are assessed with respect to their outputs: the \emph{\hvn} $\TargetNode$ and the \emph{successor represenation} $\SR$. Predictions regarding targeted \hvns $\TargetNode_{pred}$ are evaluated against ground-truth labels $\TargetNode_{true}$ using weighted F1 scores and confusion matrices. Predicted successor representations $\SR_{pred}$ are evaluated against true attack paths $\SR_{true}$ using our novel distance metric: the \mtrclong (\mtrc). Given that \mtrc scores are computed on a per-sample basis, violin and box plots are used to visualize their distribution and summary statistics over the test set.

For each prediction task, we explore the influence of both past trajectory exposure ($N_{past}$) and network size on predictive performance. To investigate how $\SR$ predictions vary along spatial and temporal dimensions, we add two further evalution conditions. First, we evaluate predicted successor representations over three discount factors $\gamma \in \{0.5, 0.95, 0.999\}$ to emphasize rewards at various points in the future ($\gamma=0.5$ represents short-term attack trajectories, while $\gamma=0.999$ represents long-term trajectories). Second, to better understand any irregularities in \gigo's predictions or \red agent behaviour, we equip the \mtrc with our novel weighting function $\Wfn$ (\autoref{eq:sr_weighting}) to evaluate how accuracy varies with a custom \emph{node remoteness} feature, defined as the length of the shortest path from a given node to the entry or targeted \hvn. This allows us to investigate if and when predicted or ground-truth attack paths venture into remote regions of the network that lie away from the path between the entry and high-value node. Weighted \mtrc scores are denoted $\srsq$. We fix $f=0.1$ wherever $\Wfn$ is used throughout this report. For brevity, since we are only using a single node feature in our weighting configuration, $\theta$ will refer to the node remoteness weighting coefficient for the remainder of the report.

For each prediction task, \gigo's performance is benchmarked against our dense-layered \tomnet variant, \gido, on the same test set. Results for each of these are summarized below.

\subsection*{How well can \gido/\gigo characterize various cyber-attacking agents?} \label{sec:experiments:agent_characterization}

\begin{figure}[h]
    \centering
    \includegraphics[width=0.9\columnwidth]{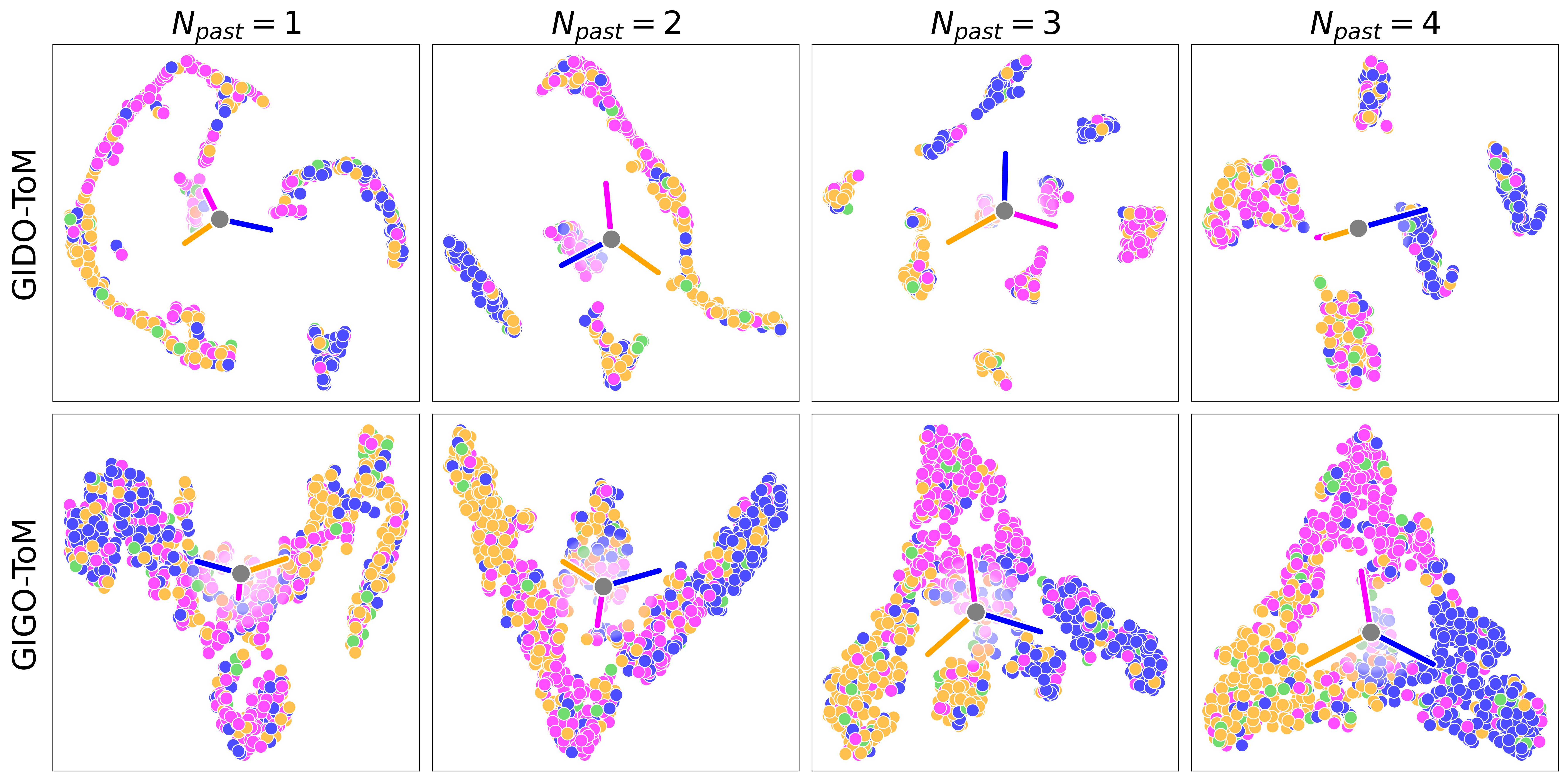}
    \caption{Test set character embeddings $\characterfull$ for 200 unseen \red agents in the \treeMixed network setting. The character network was exposed to a different number of past trajectories ($N_{past}$) for each training run. Points are colour-coded according to the agents' policies over \hvt preference. Green points represent instances where \blue prevents \red from reaching the \hvt. Colored lines are drawn from the overall data mean to the centroid of each corresponding cluster. Longer, more divergent lines therefore indicate better class separation.}
    \label{fig:exp1:char_embs}
\end{figure}

\autoref{fig:exp1:char_embs} shows clustered character network embeddings $\characterfull$ for both GIGO-ToM and our benchmark model, GIDO-ToM, over hold-out test sets of unseen \red agents when given access to varing numbers of past behaviour observations ($N_{past} \in \{1,2,3,4\}$). We find that while \gido generally produces compact and clearly separable clusters, they exhibit poor intra-cluster consistency, indicating confusion between the policies of observed agents. By contrast, \gigo exhibits strong generalization capabilities across all evaluation settings, consistently producing coherent clusters that are each predominantly characterised by one of the target users.
For \gido, performance peaks at $N_{past}=3$ and degrades again for $N_{past}=4$, heavily overlapping the clusters for two of the users. For \gigo, by contrast, the consistency, compactness, and separateness of clusters improves with higher values of $N_{past}$, demonstrating its ability to learn agents' policies and generalize eﬀectively in a
few-shot manner. We accordingly hypothesize that \gigo's performance on this task would further improve with access to more past data samples.

\subsection*{How well can \gido/\gigo predict the services targeted by cyber-attacking agents?} \label{sec:experiments:hvt_prediction}

\begin{figure}[h]
    \centering
    \subfloat[]{
        \includegraphics[height=5cm]{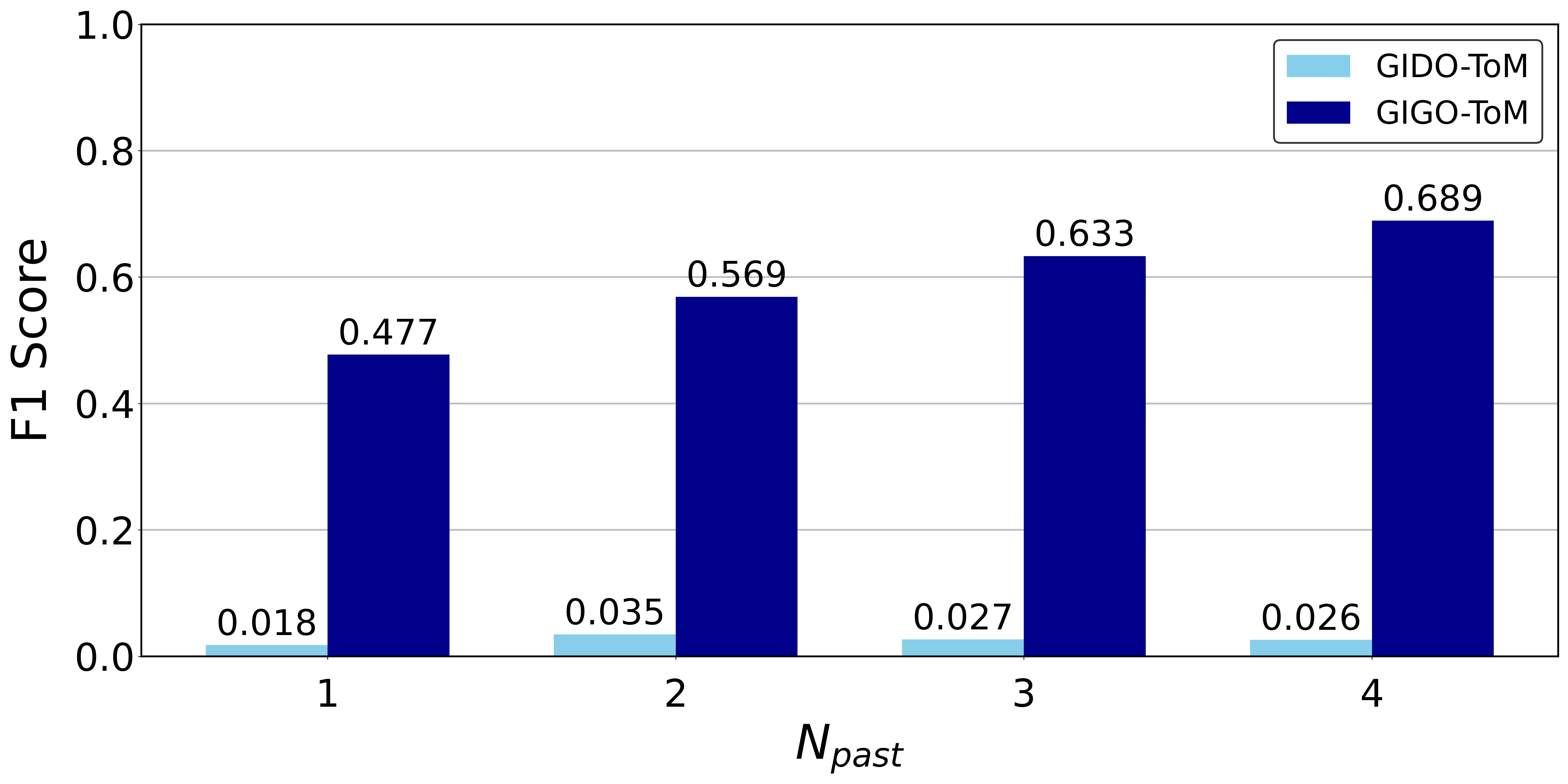}
        \label{fig:f1_scores_per_model}
    }
    \hspace{0.2cm} 
    \subfloat[]{
        \includegraphics[height=4.9cm]{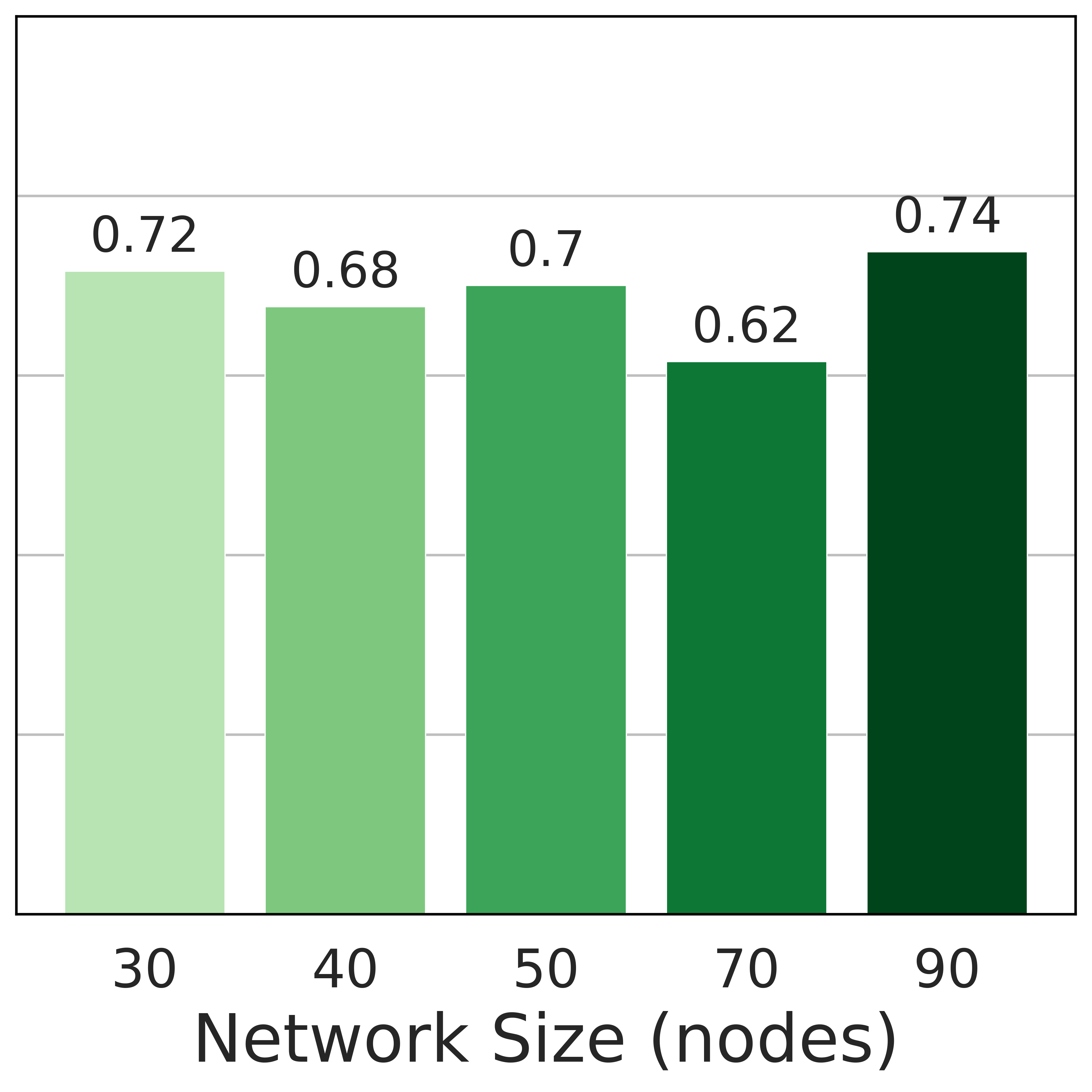}
        \label{fig:f1_scores_per_net}
    }
    \caption{
        \autoref{fig:f1_scores_per_model}: \hvn prediction F1 scores for both \gigo and \gido over the entire test set. Results are recorded for runs across different values of $N_{past}$ (number of past trajectories presented to the character network). \autoref{fig:f1_scores_per_net}: Stratified F1 scores for the \gigo, $N_{past}=4$ run, with the \treeMixed test set partitioned by network topology.
    }
\end{figure}

\begin{figure}[h]
    \centering
    \includegraphics[width=0.9\columnwidth]{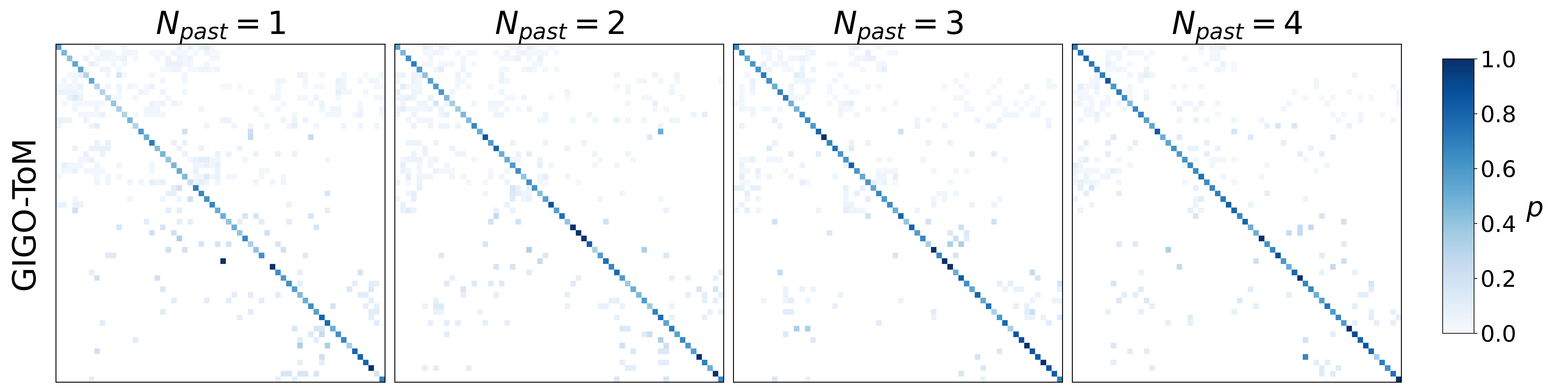}
    \caption{\gido and \gigo confusion matrices for \hvn prediction across different values of $N_{past}$ (number of past trajectories presented to the character network) for \treeMixed. For each confusion matrix, columns represent predicted target nodes and rows true target nodes. Rows are normalized so that the activation of each cell represents the proportion of predictions made for the corresponding true class ($p$).}
    \label{fig:exp2:conf_matrices}
\end{figure}

For target \hvn prediction we observe that \gigo vastly outperforms the benchmark, consistently achieving weighted F1 scores exceeding an order of magnitude higher across all evaluation settings (\autoref{fig:f1_scores_per_model}). At first glance, \gigo's peak F1 score of $0.6893$ ($N_{past}=4$) suggests moderate performance. However, given the complexity of the classification problem, with 60 possible \hvns across the \treeMixed topology, this score indicates a strong predictive capability. Such is clear from \autoref{fig:exp2:conf_matrices}, which depicts row-normalized confusion matrices over the set of all viable \hvns accross the \treeMixed topologies. The clearly distinguishable diagonals for each evaluation setting demonstrate minimal confusion between predicted target nodes.

We initially hypothesized that the class proliferation introduced by the larger networks within \treeMixed would degrade \gigo's performance on this task. However, by evaluating \gigo's performance on the individual network topologies within \treeMixed, we found no clear correlation between the number of nodes present in a network and predictive performance (\autoref{fig:f1_scores_per_net}). What did affect performance, however, was the number of branches (and therefore possible \hvt locations) present in the network. This is clear from the performance drop observed for the \treeForty and \treeSeventy topologies, which had 6 and 8 branches respectively, compared with just 4 for the others. This suggests that while \gigo's \hvt prediction capabilities are robust to high-dimensional feature spaces within the range tested in this paper, it is vulnerable to higher rates of misclassification as the number of network branches increases.

\subsection*{How well can \gido/\gigo predict the attack trajectories of cyber-attacking agents?} \label{sec:experiments:sr_prediction}



\begin{figure}[h]
    \centering
    \subfloat[]{
        \includegraphics[height=5.4cm]{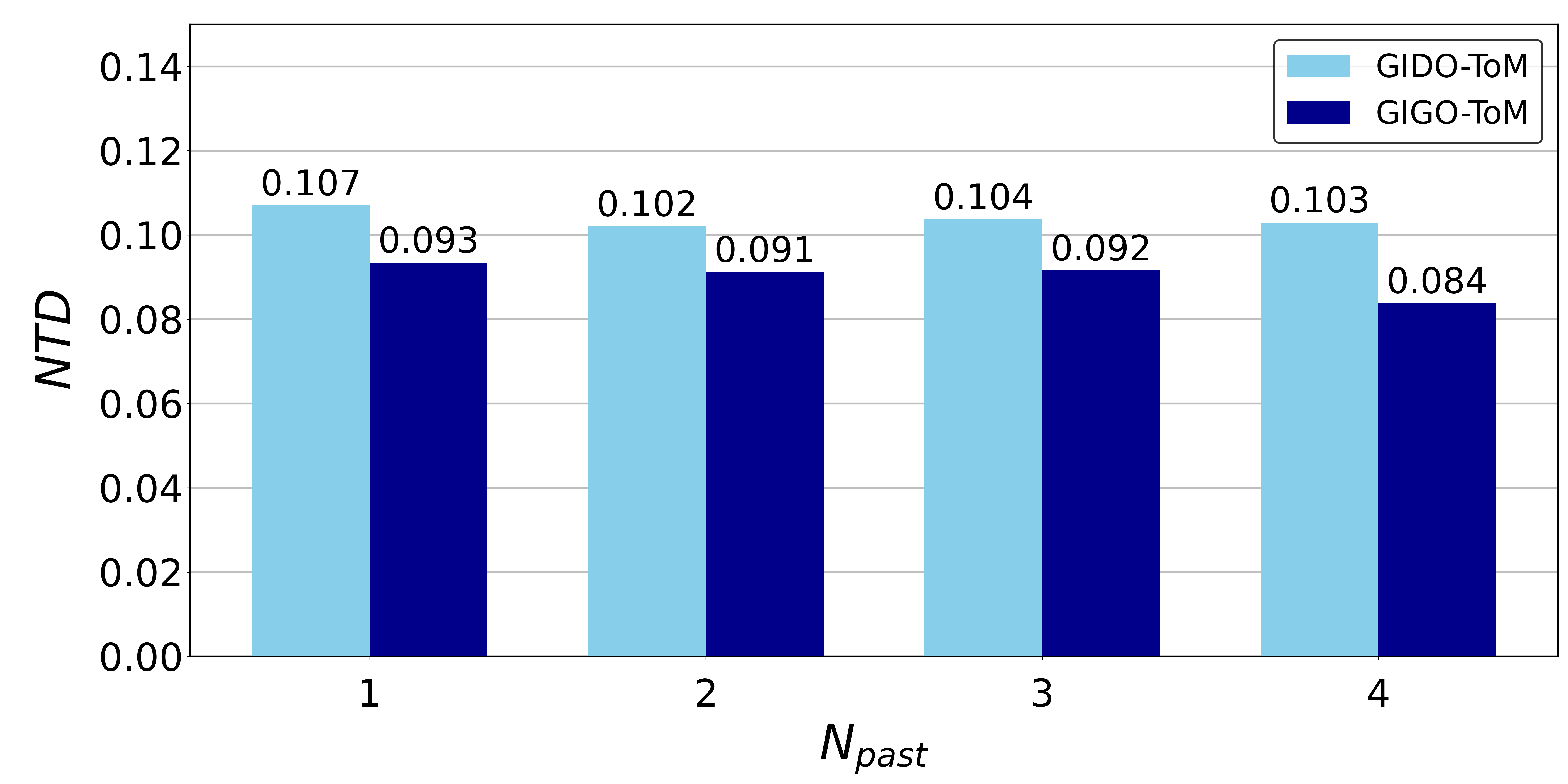}
        \label{fig:exp3:sr_bar_plot}
    }
    \hspace{0.2cm} 
    \subfloat[]{
        \includegraphics[height=5.4cm]{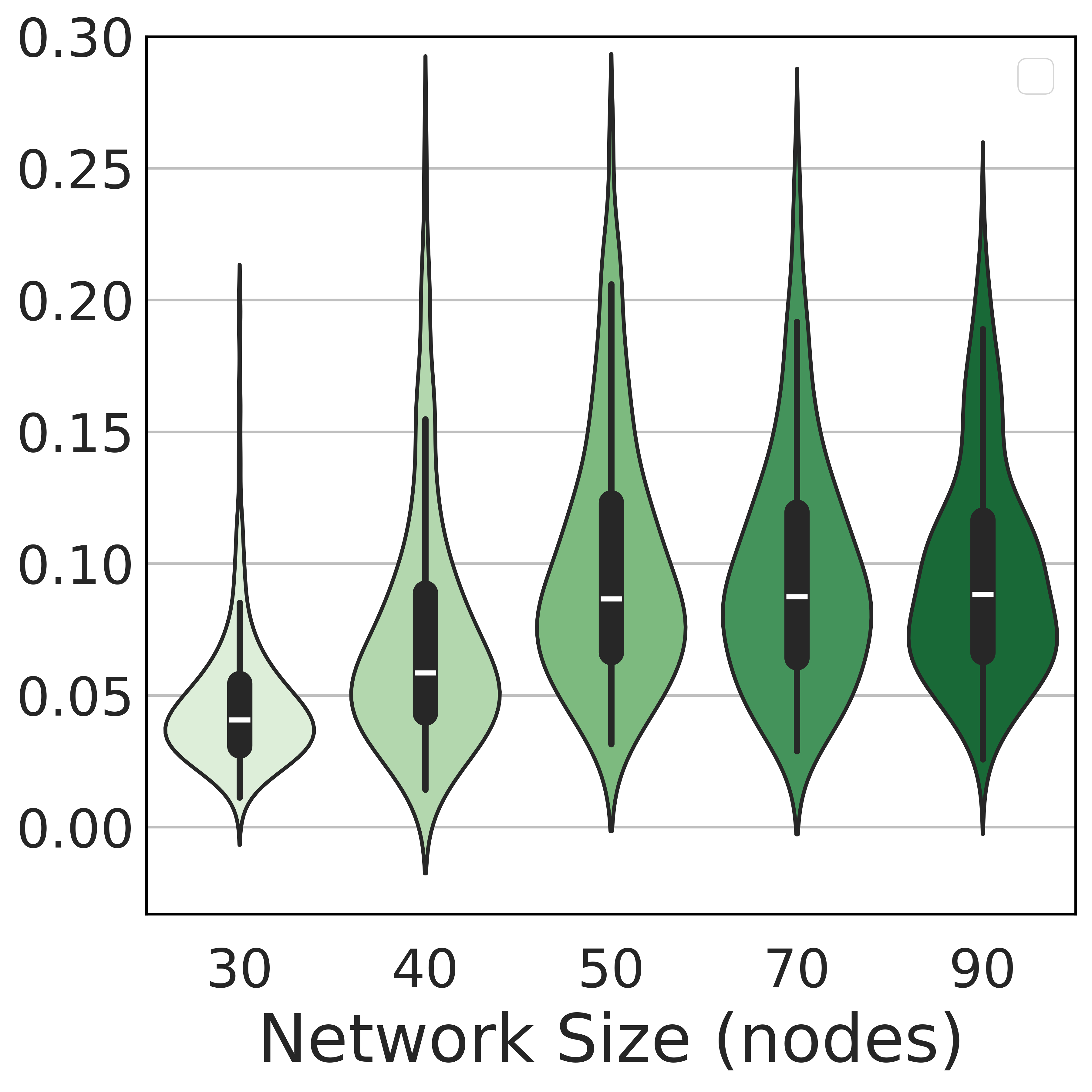}
        \label{fig:exp3:sr_violin_plot_per_net}
    }
    \caption{
        \autoref{fig:f1_scores_per_model}: Mean test set attack path prediction ($\SR_{pred}$) \mtrc scores for both \gigo and \gido across different values of $N_{past}$ (number of past trajectories presented to the character network). \autoref{fig:f1_scores_per_net}: Violin plot with overlaid box plot displaying test set \mtrc distribution and summary statistics for the \gigo, $N_{past}=4$ run, stratified by network topology.
    }
\end{figure}

In this section, we evaluate \gigo's ability to predict \red agent attack routes given a handful of past behaviour observations. \autoref{fig:exp3:sr_bar_plot} displays mean baseline (unweighted) NTD scores for each $N_{past}$ run for both \gigo and our benchmark model, \gido. We find that \gigo consistently predicts more accurate attack trajectories than \gido, with performance improving with increased past behaviour exposure, as for our previous experiments. Furthermore, we observe a negative correlation between \gigo's predictive performance and network size up to the 50-node mark, after which median \mtrc scores plateau and variability improves slightly  (\autoref{fig:exp3:sr_violin_plot_per_net}). This finding suggests that \gigo's successor representation prediction capabilities are robust to the larger networks in our test range despite having to predict progressively longer attack routes.

\begin{figure}[h]
    \centering
    \subfloat[]{
        \includegraphics[height=5.4cm]{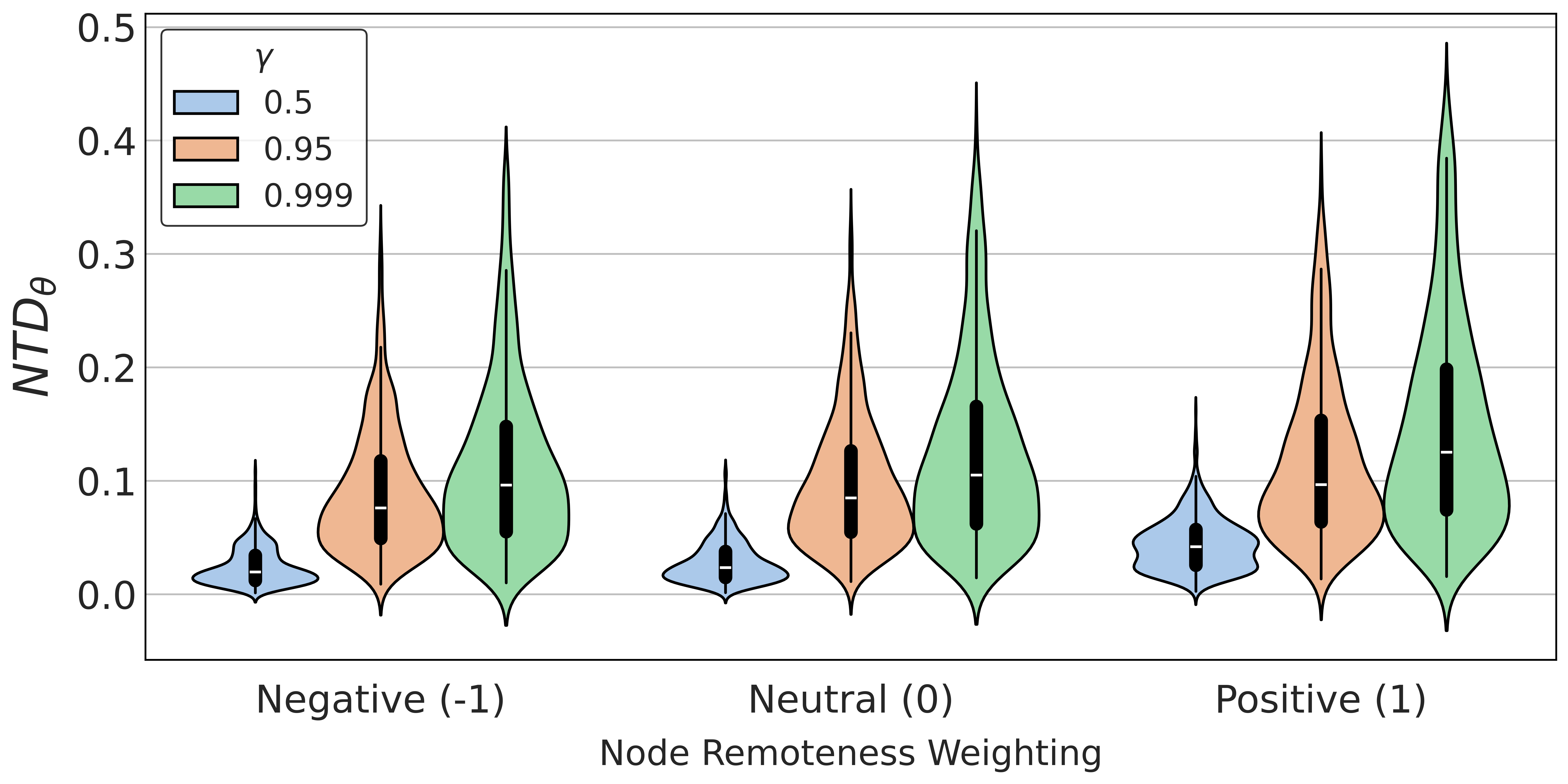}
        \label{fig:exp3:sr_gamma_c0_violin_plot}
    }
    \hspace{0.2cm} 
    \subfloat[]{
        \includegraphics[height=5.4cm]{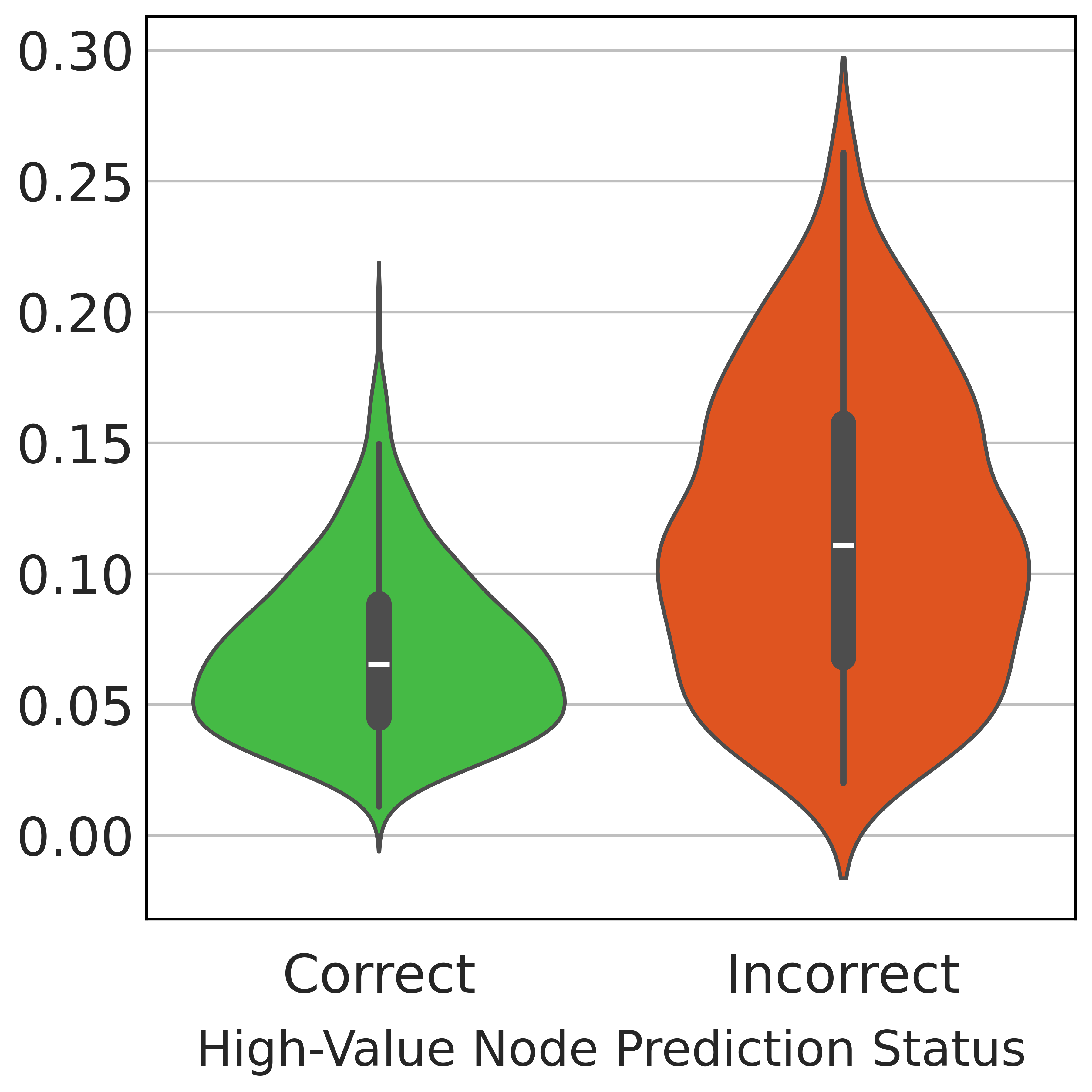}
        \label{fig:exp3:sr_hvt_violin_plot}
    }
    \caption{
        \autoref{fig:exp3:sr_gamma_c0_violin_plot}: Violin plot with overlaid box plot displaying our spatio-temporal evaluation of predicted test set attack paths: Weighted $\srsq$ scores (\gigo, \treeMixed, $N_{past}=4$), stratified by node remoteness weighting (spatial) and discount factor $\gamma$ (temporal). `Neutral' represents the baseline/unweighted \mtrc. \autoref{fig:exp3:sr_hvt_violin_plot}: Test set $\mtrc$ scores stratified by \hvn prediction status.
    }
\end{figure}

To investigate how NTD scores vary across temporal and spatial dimensions, we plot the distribution and summary statistics of $\srsq$ scores across different values of $\gamma$ (predictive time horizon) and different values of the coefficient of our \emph{node remoteness} feature (spatial proximity to the entry or targeted \hvn in graph space) in our metric weighting function $\Wfn$ in \autoref{fig:exp3:sr_gamma_c0_violin_plot}.

We firstly find that performance is significantly affected by one's choice of $\gamma$. Unsurprisingly, as before, short-term predictions ($\gamma=0.5$) are best. This is due to the relative complexity of the \yt environment, which presents a significantly more challenging learning problem when the model is forced to account for more distant future states. The same is true spatially, underscored by the significant $\srsq$ deterioration observed when inverting the node remoteness weighting from -1 to 1 (thereby switching the emphasis from nodes closer to the \red agent's start and end points to those further away).

This could indicate one of two possibilities: either \gigo predicts paths into remote regions of the network that the \red agent does not in fact visit, or the \red agent visits remote regions of the network that \gigo fails to predict.
One would expect to see instances of the latter if the \red agent were often pushed off-course by the \blue agent's defensive activity. However, given the strong determinstic policies of the \redHVTPreferenceSP agents used here and our inclusion of only episodes that result in \red agent victory, the former is more likely.

It should be noted, however, that, this does not necessarily entail that \gigo has simply predicted a single path to the wrong \hvn. Indeed, we do observe that \gigo's predicted attack paths deviate significantly further from the ground-truth for samples where the incorrect target \hvn is predicted, confirming that the two outputs are largely aligned (\autoref{fig:exp3:sr_hvt_violin_plot}). However, the interesting predictive idiosyncrasies that are possible here warrant further investigation.

\begin{figure}[h]
    \centering
    \includegraphics[width=0.85\columnwidth]{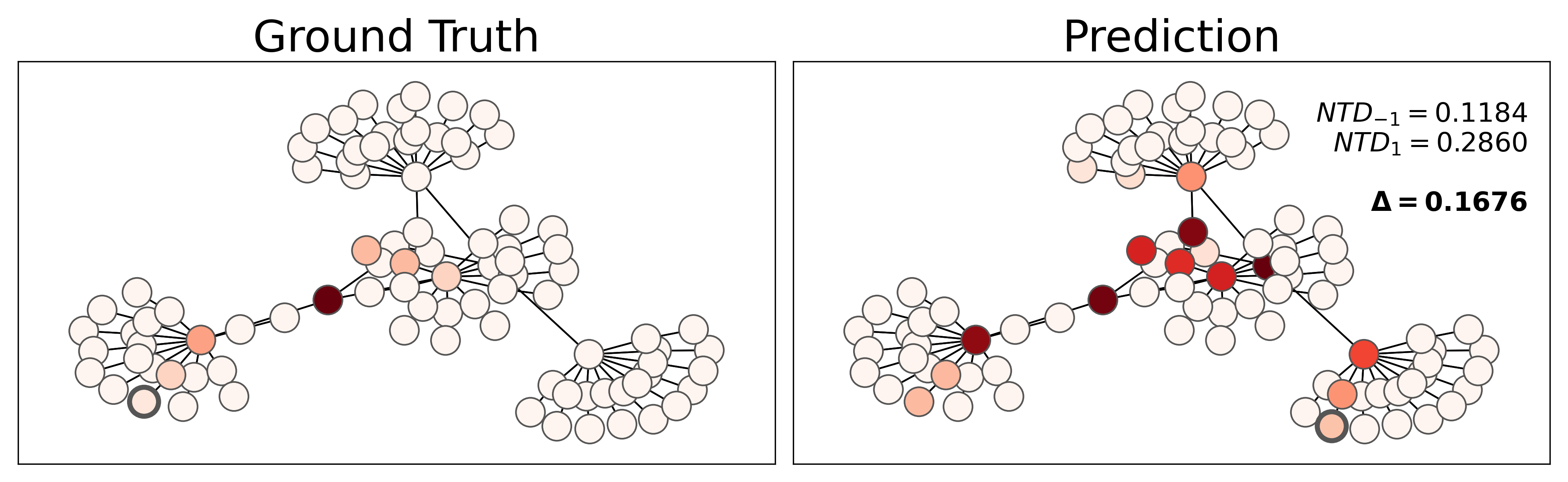}
    \caption{Evidence of \gigo's `hedging' behaviour: $\SR_{true}$ vs. $\SR_{pred}$ for the test sample with the largest discrepancy between $NTD_{-1}$ and $NTD_{1}$ (i.e. for which the node remoteness weighting produces the greatest difference in $\srsq$ score). $\TargetNode_{true}$ and $\TargetNode_{pred}$ are emboldened in each plot, respectively. To challenge \gigo, only samples from the most challenging evaluation sub-setting (\treeNinety, $\gamma=0.999$) were considered here. $\Delta$ represents the absolute difference between the scores produced by the positive and negative node remoteness weighting coefficient. Displayed in this plot is the largest encountered over the test subset. We observe that \gigo predicts paths to multiple \hvns.}
    \label{fig:exp3:max_diff_SR}
\end{figure}

As a principled next step, we inspect ground-truth vs. predicted successor representations for the test sample with the widest \mtrc range spanned across our node remoteness weighting settings (\autoref{fig:exp3:max_diff_SR}). This will reveal, in simple terms, the sample for which \gigo's predictive accuracy varies most with respect to proximity to the entry and preferred
\hvn. To challenge \gigo, we consider only samples for the hardest evaluation setting within the \treeMixed, $N_{past}=4$ run: \treeNinety, $\gamma=0.999$.

\begin{figure}[h]
    \centering
    \includegraphics[width=0.65\columnwidth]{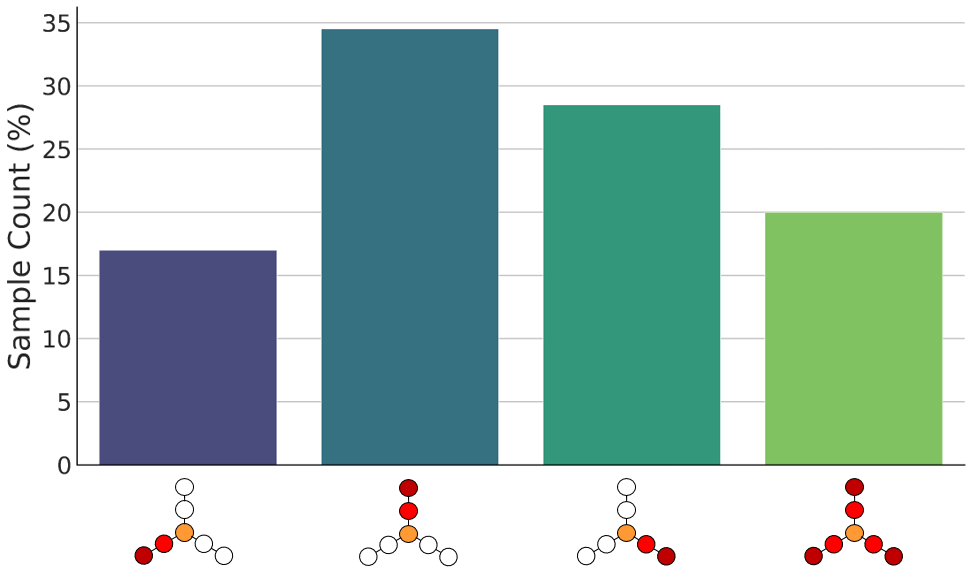}
    \caption{Histogram displaying the distribution of test samples for the \treeNinety, $\gamma=0.999$ condition across different k-means groupings. K-means over the predicted successor representations for these samples with $k=4$ (3 \hvns plus a `miscellaneous' bin) exposed groupings based on approximate attack path. The first three bins correspond to determinate attack paths to a single network branch, and therefore represent `confident' predictions. The final bin contains samples for which no single path was discernible. We hypothesize that hedged predictions result from observing a set of pat trajectories where multiple users were consumed by the Red agent.}
    \label{fig:exp3:kmeans_histogram}
\end{figure}

\autoref{fig:exp3:max_diff_SR} reveals interesting behaviour. Specifically, \gigo seems to hedge it's bets with respect to the \red agent's attack path, mapping routes from the entry node to every \hvn rather than comitting to any single one.
Indeterminate attack path predictions have little utility to network operators. Therefore, to investigate the prevalence of these non-committal predictions in our test set, we ran K-means over the predicted successor representations for these samples with $k=4$ (3 high-value nodes plus a `miscellaneous' bin). This exposed groupings based on approximate attack path, revealing that these indeterminate predictions constitute roughly 20\% of the those made over the test set for \treeNinety, $\gamma=0.999$ (\autoref{fig:exp3:kmeans_histogram}).


In summary, \gigo predicts attack paths to a level of accuracy that is certainly useful from a cyber-defence perspective. Even under our most challenging evaluation settings, distinctive attack trajectories are predicted for approximately 80\% of test samples, averaging a mean \mtrc of 0.08. This performance is robust to high-dimensional network spaces, and improves with exposure to more past behaviour observations.


%% file: sections/discussion.tex
\section*{Discussion}

In this work, we evaluate the efficacy of \tomfull-inspired metacognitive model architectures in providing actionable insights into the goals and behavioural patterns of adversarial cyber agents.
We find that, through observing past episodes of \blue
and \red cyber agents interacting with each other within the \yt environment,
our graph-based \mtomfull approach, \gigo, can learn character embeddings that can be clustered based
on the \red agent's preferred outcome, thereby providing a means to characterize observed offensive cyber agents.
We also find that our architecture can accurately predict \red's preference over
\hvns, as well as the intermediate nodes that are
likely to come under attack as the episode unfolds.

We note that there are numerous avenues for future research in this area,
including a number of experiments that one
could run with our framework.
For example, we are keen to design representative `Sally Anne'
test\cite{wimmer1983beliefs,baron1985does} scenarios for cyber-defence to evaluate the extent
to which \gigo can attribute false beliefs to observed agents. This would serve the dual purpose of both stress-testing the mental network, which was deactivated for our experiments, and conclusively determining whether \gigo makes its predictions based on a true theory of mind rather than learned shortcuts \cite{geirhos2020shortcut}.

\tom models are also often used as opponent models\cite{piazza2023limitations},
raising questions as to whether \gigo could be extended to provide an
explicit belief model for partially observable environments with
a graph-based observation space.

Finally, we consider that our novel similarity measure, the \mtrclong (\mtrc), holds potential for the direct optimization as well as evaluation of predicted successor representations. The computatuon of the Wasserstein Distance (on which the \mtrc is based) requires the resolution of a linear program, and is therefore  both computationally intensive and non-differentiable, precluding its use as a loss function. However, differentiable approximations exist. The Sinkhorn Distance, for example, introduces an entropic regularization term to the optimal transport problem that smooths the objective function, permitting well-defined gradients that can be computed efficiently\cite{sinkhorn1967concerning}. This raises the possibility of repurposing the \mtrclong as a loss function against which predicted graph-based successor representation can be optimized directly.
To broach this topic, we developed a Sinkhorn-based \mtrc loss function and ran the preliminary test of comparing evaluation/test set \mtrc scores produced by this loss and the soft label cross-entropy loss when optimizing \gigo \emph{only} with respect to successor representation predictions (i.e. dropping $\Lhvt$ from \autoref{eq:total_loss}). Our initial findings suggest that the \mtrc may significantly improve predicted successor representations when used as a loss function. Preliminary results obtained using this loss function are provided in \autoref{app:ntd}.

In summary, our work shows that GNN-based \tomnet architectures show promise for the field of cyber defence. Under the abstract experimental conditions of this paper, the breadth and accuracy of \gigo's predictions are of a sufficient caliber to be leveraged by cyber-defence specialists to anticipate the goals and behaviours of observed cyber attackers in real time, enabling focused and timely defensive measures. We deem these findings compelling enough to warrant further investigation, and it is our hope that they are sufficient to inspire others to carry this work forward to more challenging and realistic cyber defence scenarios.

We hope to have also demonstrated that our graph-based similarity metric, the \mtrclong, provides a more flexible, intuitive, and interpretable means of comparing graph-based probability distributions than anything found during our review of existing literature. Beyond the obvious merits of a bounded and network-agnostic evaluation measure, our weighting method enables creative strategies for fine-tuning one's understanding of network vulnerabilities and attacker behaviour. We hold that these constitute valuable additions to the network analysis toolbox, with potential applications beyond the realm of cyber defence.

We hope that this work will raise awareness regarding the potential benefits
of graph-based \tom for cyber-defence and that our work will inspire readers
to attempt to answer some of the research questions that we have raised.

%% file: appendix/networks.tex
\section{Additional Custom \yt Network Topologies} 
\begin{figure}[H]
    \centering
    \subfloat[\forest]{
    \label{fig:ForestNetwork}
    \includegraphics[width=0.5\columnwidth]{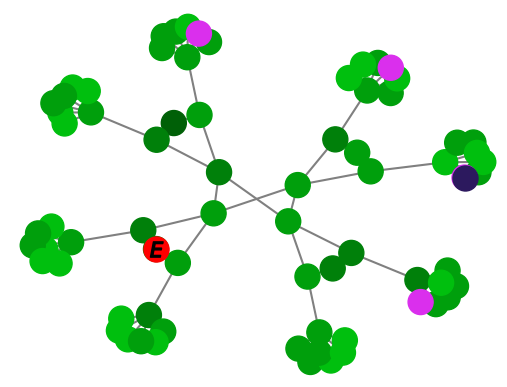}}
    \subfloat[\oc]{
    \label{fig:oc}
    \includegraphics[width=0.5\columnwidth]{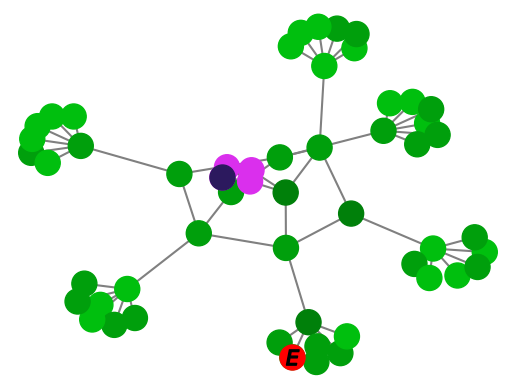}}
    \caption{Additional \yt Network Topologies}
    \label{fig:AdditionalNetworkTopologies}
\end{figure}

In addition to the custom network topologies described in the \hyperref[sec:problem_forumulation:gyt:networks]{Preliminaries} section, we developed two more complex ones for more rigorous future testing:

\begin{itemize}
    \item \forest: A network that resembles a forest layout, consisting of 72 nodes, shown in \autoref{fig:ForestNetwork}.
    \item \oc: A network consisting of 54 nodes, where in the default setting
    the high value nodes are situated on one of our four servers at the centre of the network depicted in \autoref{fig:oc}.
\end{itemize}

%% file: appendix/agents.tex
\section{Rule-Based Blue Cyber-Defence and Red Attacking Agents} \label{app:agents}

In addition to the agents used for the experiments in this report, we initially developed a suite of other \blue cyber-defence and \red cyber-attacking agents for our evaluations.
We first provide an overview of the actions available to each agent,
before outlining the agents that we implemented.
\noindent The goal of the \blue agent is to stop the \red agent's attack and survive for 500 time-steps. The actions that the \blue cyber-defence agents can take to this end are:
\begin{itemize}
    \item \donothing: performs no action.
    \item \scan: reveals any hidden compromised nodes.
    \item \msn: removes the infection from a node.
    \item \rnv: lowers the vulnerability score of that node.
    \item \restore: uses \msn and resets the node vulnerability at the start of the episode.
    \item \isolate: removes all connections from a connected node.
    \item \reconnect: returns all connections from an isolated node.
\end{itemize}

\noindent The goal of the \red cyber-attacking agent, on the other hand, is to evade the \blue agent in order to capture a high-value node within 500 time-steps, starting from a specific entry node. \red has its own set of offensive actions, including:
\begin{itemize}
    \item \donothing: performs no action.
    \item \basicattack: attempts to infect a node.
    \item \randommove: moves the red agent to another connected node.
    \item \zda: guaranteed infection on a node. Red agents have a \zda budget, gaining an additional \zda at a defined rate per time-step.
    \item \spread: attempts to infect all nodes connected to any compromised node.
    \item \intrude: attempts to infect all nodes.
\end{itemize}

\medskip

\noindent For our \blue agents, we have:

\begin{itemize}
    \item \blueSleep: Takes the \donothing action at every step, used to benchmark the red agent's performance as if there were no blue agent present on the network
    \item \blueRandom: Selects a random action from the action-space, and then a random node if that action requires a target. This is used to benchmark against a completely random blue policy with no defence strategy.
    \item \blueRandomSmart: Selects a random action from the entire action-space, but chooses an appropriate node for that action.
                       For example, it will only use \msn and \restore on nodes that the agent can observe are compromised.
                        This is used to observe how a blue agent would perform if it had no strategy in regards to its action selection, but made logical choices for its targets.
    \item \blueIsolate: Takes the \isolate action on the entry and high-value nodes at the start of the episode, cutting all possible routes the red agent has to the high value nodes.
                        It will then take \scan for the rest of the episode until a new compromised node is revealed, where it will then take \msn on the compromised nodes.
                        This exploits the rules of the game for a guaranteed blue victory, but can be used to benchmark the score an agent would get through following this scorched-earth policy.
\end{itemize}
We have six agents that utilise the \msn or \restore actions more intelligently to mitigate the attack.
Each of these operate by selecting a node based on its distance from any high-value node,
assigning greater priority to compromised nodes that are close to a high-value node:
\begin{itemize}
    \item \blueMSND: A deterministic agent that takes the \msn action if a compromised node is within three hops of any high value node, otherwise takes \scan.
                        The strategy here is to protect only the core of the network. \blue is content with sacrificing the edge nodes to do so.
    \item \blueMSNS: A stochastic agent. Takes \msn and \scan based on a probability. The \msn probability increases if a compromised node is close to a high-value node.
                        This agent's strategy is to protect the entire network, but will look to remove infections closer to the high value nodes with greater urgency.
    \item \blueRestore: The same as \blueMSNS, but taking \restore rather than \msn to observe the difference in performance between the \msn and \restore actions.
    \item \blueMSNRNV: The same as \blueMSNS, but will uniformlyselect between \scan or \rnv of the most vulnerable node if not taking \msn.
                        This is to assess if it is better for the agent to prioritise monitoring the network, or to make the network more resilient to attacks.
    \item \blueMSNRestore: The same as \blueMSNS, but will randomly select \msn or \restore if not taking \scan.
                        This can be used to train learning agents under what conditions it would be better for the blue agent to take \msn or \restore.
    \item \blueMSNRNVRestore: This agent uniformly selects between \msn and \restore when removing infections, and between \scan and \rnv otherwise to detect and protect against red activity respectively.
                        As with the other stochastic agents, the probability that it selects between these two sets of actions corresponds to the distance the nearest compromised node is to a high value node.
                        This allows the agent to sample the full range of defensive capabilities, excluding the \isolate action which was shown to be too powerful and restrictive as an action.
                        {Combines \msn\ -- \rnv and \msn\ -- \restore to take all four actions based on the action probabilities.}
\end{itemize}

It is worth noting that only the \blueRandom and \blueRandomSmart take the \reconnect action.
This is because the action is only useful to reconnect isolated nodes, and no other agents take the \isolate action due to its power relative to the other defensive actions.
Using \reconnect can improve the agent's score as it is punished for each isolated node in the reward function, but is redundant without \isolate.

\medskip

\noindent For our cyber-attacking agents, we seven \red agents with action probabilities sampled from a Dirichlet distribution $\pi \sim Dir(\alpha)$:
\begin{itemize}
 \item \redRandomSimple: Selects an action weighted by above action probabilities and selects random valid targets. This
                            agent simulates an unskilled attacker with no prior knowledge or plan of attack. It occasionally wastes time trying to use \zda, even when unavailable.
 \item \redRandomSmart: First selects an action and target exactly like \redRandomSimple but if \zda is chosen and, if none are available, uses a \basicattack instead. This
                            is a more proficient version of \redRandomSimple.
 \item \redTargetConnected: Selects an action exactly like \redRandomSmart but always selects the target with most connections. This
                            agent uses a disruptive attack strategy as its targets are highly connected, trying to reveal as much of the network
                            as possible.
 \item \redTargetUnconnected: Selects an action exactly like \redRandomSmart but always selects the target with the least connections. Using
                            the opposite strategy to \redTargetConnected, this agent makes the assumption that the most valuable data would be stored on the most secluded nodes.
 \item \redTargetVulnerable: Selects an action exactly like \redRandomSmart but always selects the target with highest vulnerability. The
                            strategy of this agent is to compromise as many nodes as quickly as possible because the more vulnerable the target, the more
                            likely the attack is to succeed.
 \item \redTargetResilient: Selects an action exactly like \redRandomSmart but always selects the target with lowest vulnerability. Using
                            the opposite strategy to \redTargetVulnerable, this agent perhaps makes the assumption that the most valuable data would
                            be stored on the most resilient nodes.
 \item \redHVTSimple: Behaves exactly like \redRandomSimple until a high value node becomes connected to a compromised node,
                      at which point actions and target choices become deterministic -- \zda if available, else \basicattack on a high value node. This
                      agent still has no prior knowledge of the network but knows the importance of compromising the high-value node.
\end{itemize}

There are two red agents with high value node preferences, instead of action probabilities, sampled from a Dirichlet distribution:

\begin{itemize}
\item \redHVTPreference: Selects a particular high value node to aim for at the start of the episode. The selected high-value node is based on the high value node preferences as well as the shortest distance between high value nodes and entry nodes. This
                         attacker has prior knowledge of the network structure, for example insider information, meaning it can calculate path lengths and is then likely to take the most direct path to its chosen high-value node.
\item \redHVTPreferenceSP: Selects a particular high value node to aim for, exactly like \redHVTPreference, but always takes the shortest path. This
                            makes it a more ruthless version of \redHVTPreference.
\end{itemize}


\subsection*{Rule-Based Agent Evaluation} \label{app:agents:eval}

Upon implementing the agents described in the above section we conducted
an extensive evaluation, allowing us to answer the question:
`What are the optimal mixtures over our respective sets of agents
that Red and Blue can sample from for different game settings?'.
Here, we conduct an empirical evaluation using three network topologies:
\treeThirty; \oc and \forest.
Each network is implemented with three entry nodes.
As always, \hvns are situated
on leaf nodes.
We conduct 100 evaluation episodes for each \blue, \red joint-policy
profile, and subsequently evaluate the agents with respect to
\blue rewards, \blue win rate, and episode duration.
For the latter, a 500 step episode represents a \blue victory.

We provide detailed tables with the results from each of the
selected topologies in tables \ref{table:ForestNetwork:durations} -- \ref{table:OpticalCoreNetwork:results}.
Tables \ref{table:all:rewards} -- \ref{table:all:durations} meanwhile
provide the averages.
It should not come as a surprise that the strongest \blue cyber-defence
agent with respect to win rate and duration is \blueIsolate
(See Tables \ref{table:all:results} and \ref{table:all:durations}).
However, isolating nodes comes at a cost with respect to the \blue
reward (See \autoref{table:all:rewards}).
When discarding \blueIsolate it becomes apparent that the \red cyber-attacking
\redHVTPreferenceSP presents a particular challenge to the remaining \blue
cyber-defence agents.
Blue's best response against \redHVTPreferenceSP is \blueRandomSmart with respect
to average duration (140 steps, see \autoref{table:all:durations}) and win
rate (0.21, see \autoref{table:all:results}).
However, due to selecting the isolate, \blueRandomSmart performs poorly
with respect to the \blue reward function (-942.48, see \autoref{table:all:rewards}).
In contrast \blueMSND achieves the best result with respect to \blue rewards (-58.37)
against \redHVTPreferenceSP, representing the pure strategy Nash equilibrium within
the \blue rewards payoff table.
Other strong performers against \redHVTPreferenceSP with respect to the reward
criteria include \blueMSNS (-66.23), \blueMSNRNV (-69.13), \blueMSNRNVRestore~(-78.6)
and \blueMSNRestore (-78.01).
\begin{table}[H]
\centering
\includegraphics[width=1.\columnwidth]{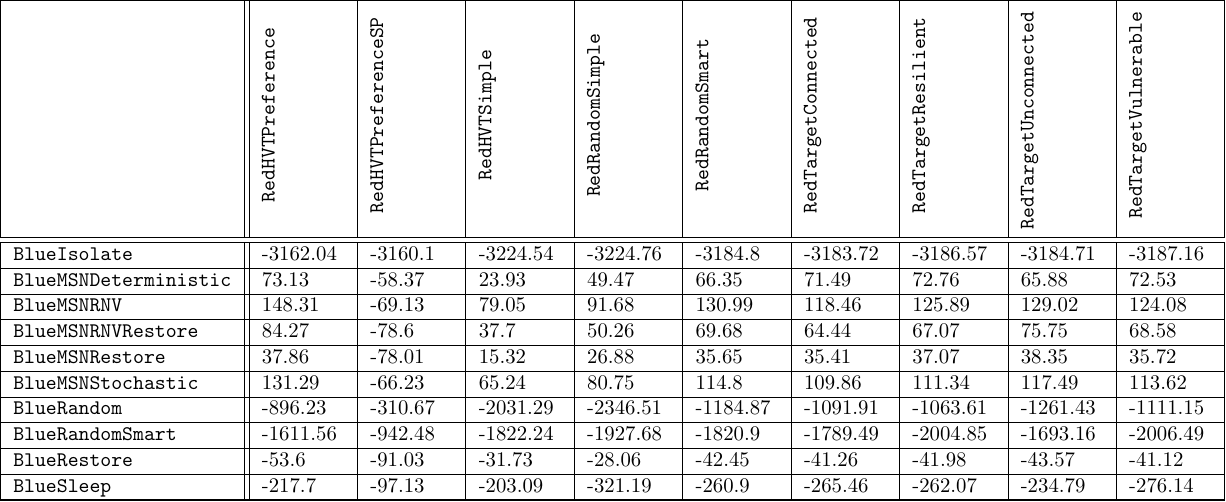}
\caption{Rewards for deterministic red agents ($\alpha=0.01$). Averaged across the three networks.}
\label{table:all:rewards}
\end{table}
\begin{table}[H]
\centering
\includegraphics[width=1.\columnwidth]{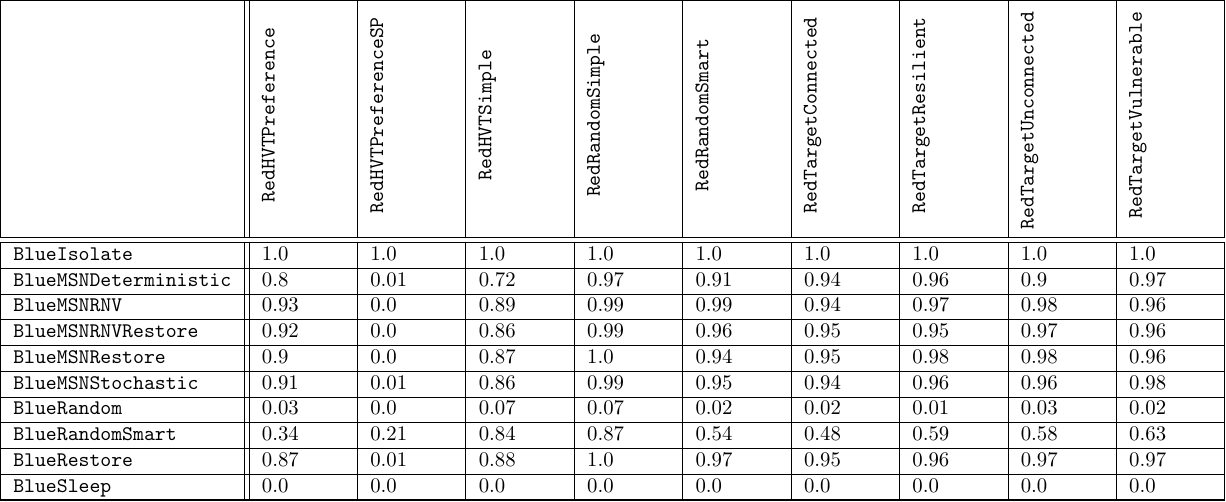}
\caption{Win rates for deterministic red agents ($\alpha=0.01$). Averaged across the three networks.}
\label{table:all:results}
\end{table}
\begin{table}[H]
\centering
\includegraphics[width=0.9\columnwidth]{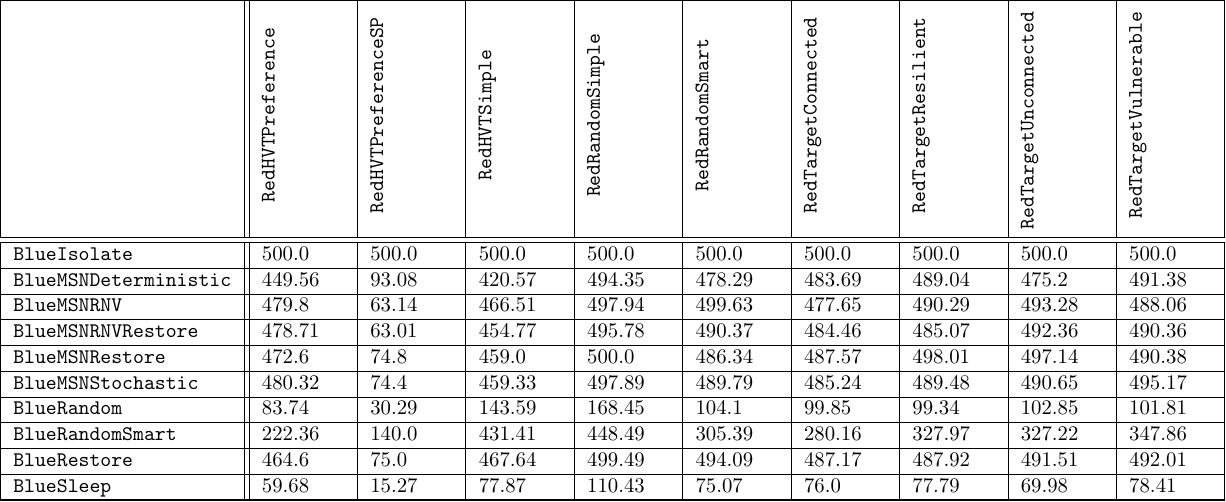}
\caption{Durations for deterministic red agents ($\alpha=0.01$). Averaged across the three networks.}
\label{table:all:durations}
\end{table}

\begin{table}[H]
    \centering
    \includegraphics[width=0.9\columnwidth]{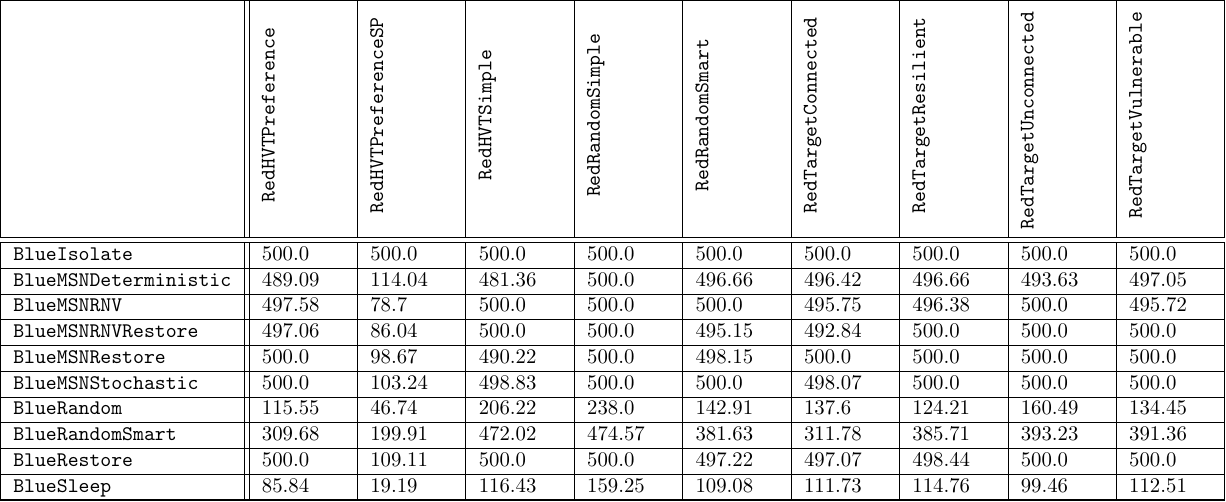}
    \caption{\texttt{ForestNetwork} Durations for deterministic red agents ($\alpha=0.01$).}
    \label{table:ForestNetwork:durations}
\end{table}

\begin{table}[H]
    \centering
    \includegraphics[width=0.9\columnwidth]{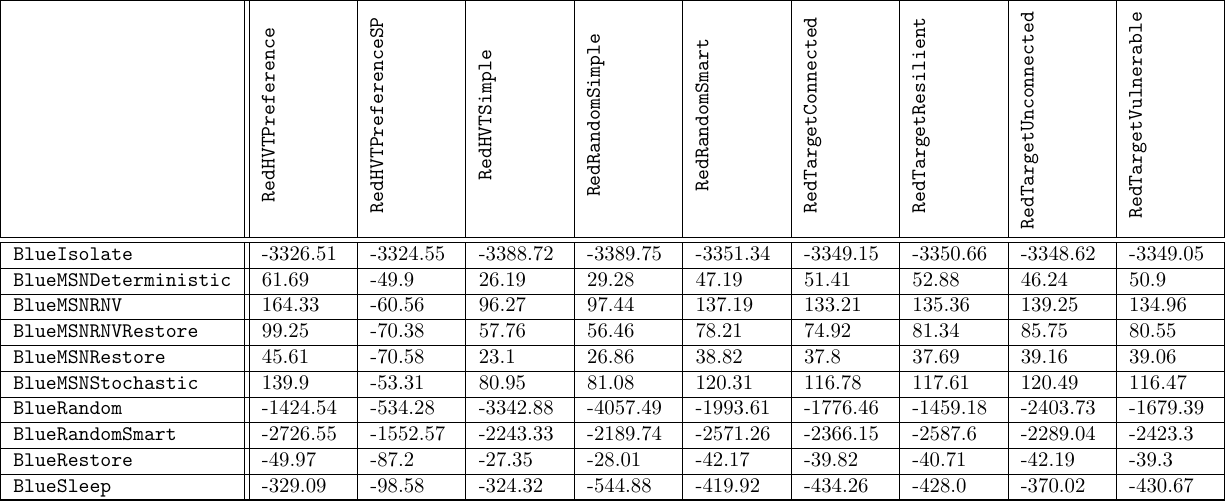}
    \caption{\texttt{ForestNetwork} Rewards for deterministic red agents ($\alpha=0.01$).}
    \label{table:ForestNetwork:rewards}
\end{table}

\begin{table}[H]
    \centering
    \includegraphics[width=0.9\columnwidth]{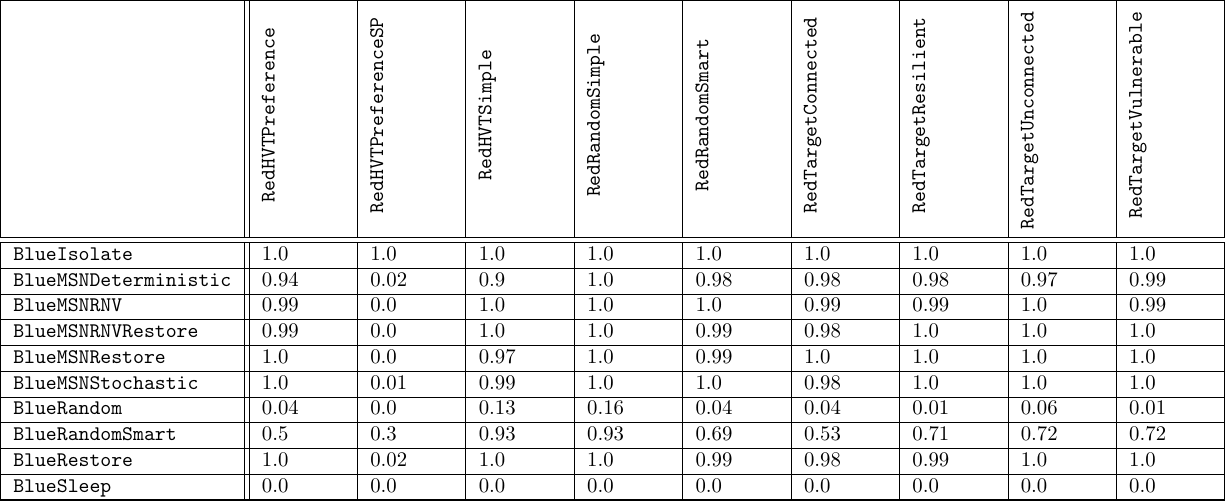}
    \caption{\texttt{ForestNetwork} Results for deterministic red agents ($\alpha=0.01$).}
    \label{table:ForestNetwork:results}
\end{table}

\begin{table}[H]
    \centering
    \includegraphics[width=0.9\columnwidth]{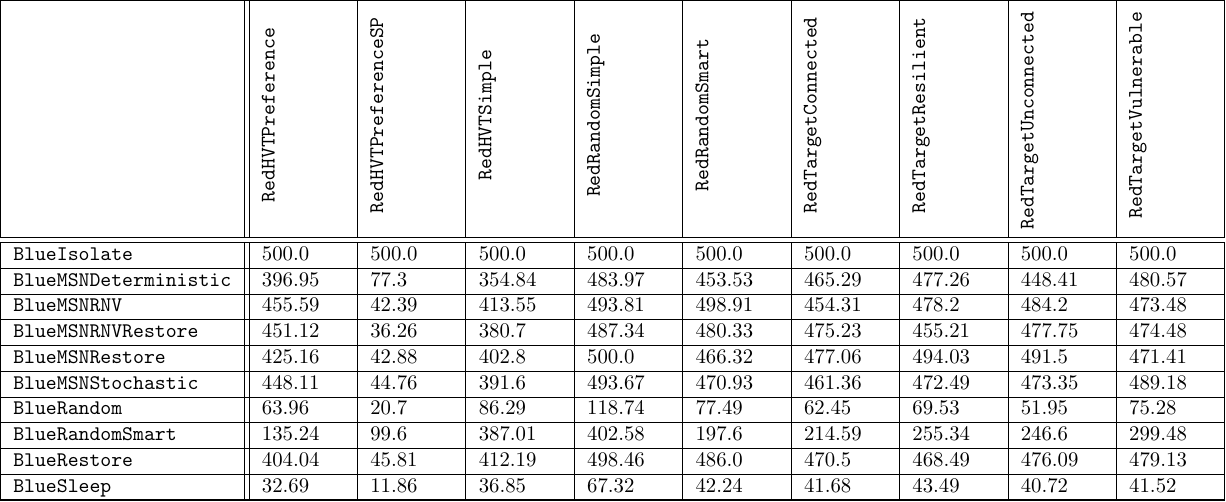}
    \caption{\texttt{TreeNetwork} Durations for deterministic red agents ($\alpha=0.01$).}
    \label{table:TreeNetwork:durations}
\end{table}

\begin{table}[H]
    \centering
    \includegraphics[width=0.9\columnwidth]{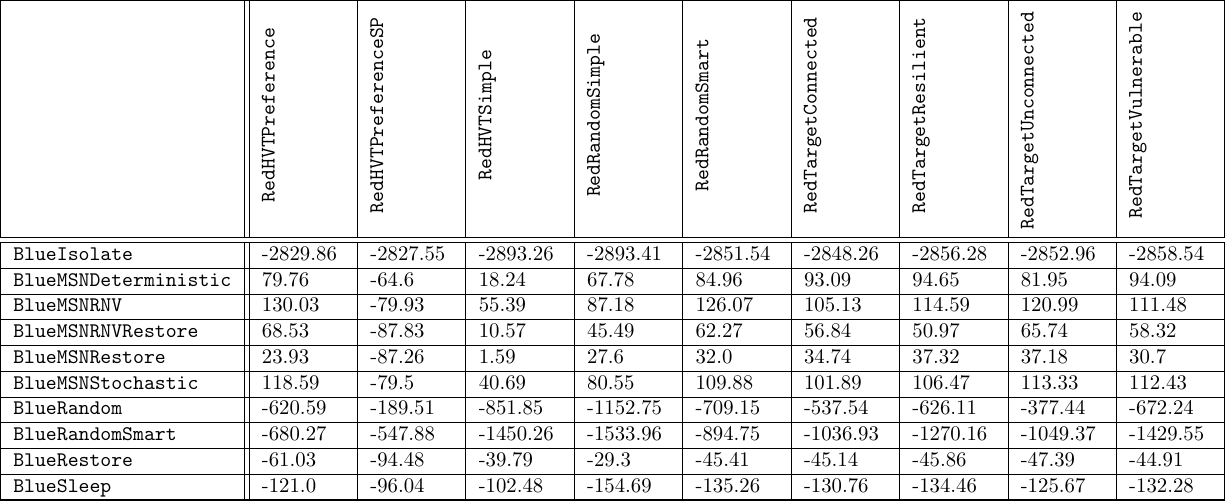}
    \caption{\texttt{TreeNetwork} Rewards for deterministic red agents ($\alpha=0.01$).}
    \label{table:TreeNetwork:rewards}
\end{table}

\begin{table}[H]
    \centering
    \includegraphics[width=0.9\columnwidth]{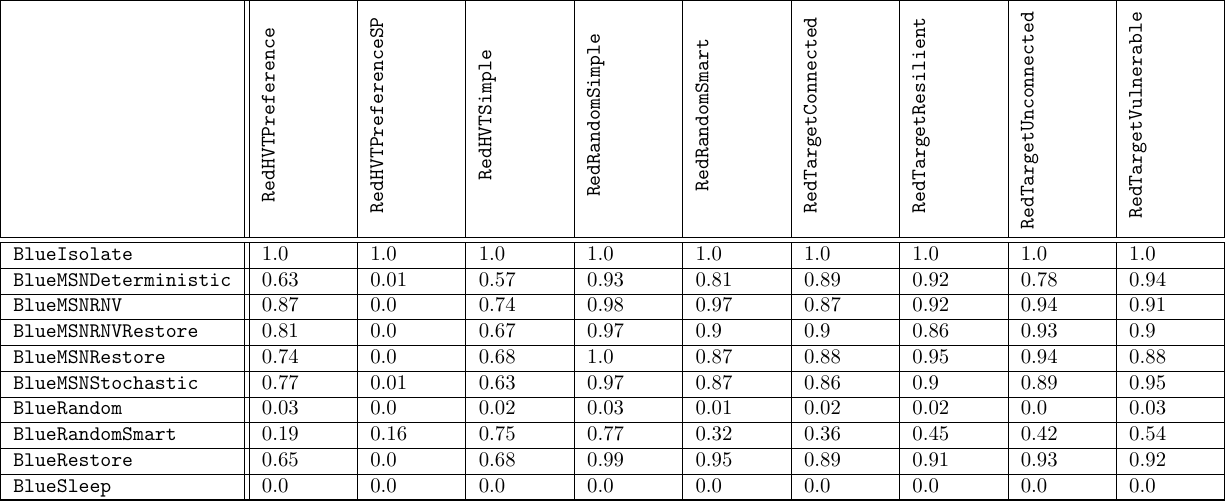}
    \caption{\texttt{TreeNetwork} Results for deterministic red agents ($\alpha=0.01$).}
    \label{table:TreeNetwork:results}
\end{table}

\begin{table}[H]
    \centering
    \includegraphics[width=0.9\columnwidth]{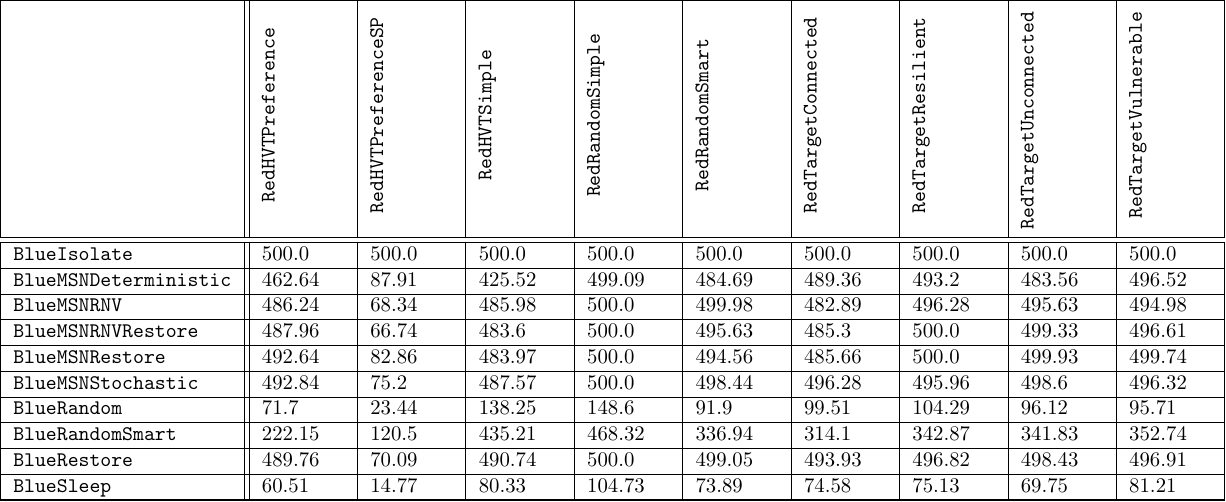}
    \caption{\texttt{OpticalCoreNetwork} Durations for deterministic red agents ($\alpha=0.01$).}
    \label{table:OpticalCoreNetwork:durations}
\end{table}

\begin{table}[H]
    \centering
    \includegraphics[width=0.9\columnwidth]{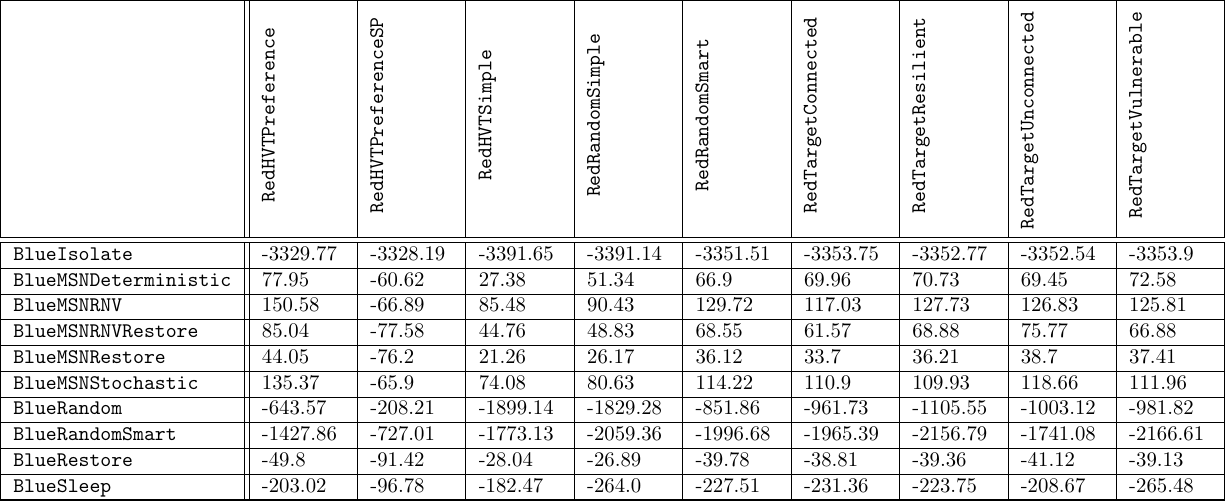}
    \caption{\texttt{OpticalCoreNetwork} Rewards for deterministic red agents ($\alpha=0.01$).}
    \label{table;OpticalCoreNetwork:rewards}
\end{table}

\begin{table}[H]
    \centering
    \includegraphics[width=0.9\columnwidth]{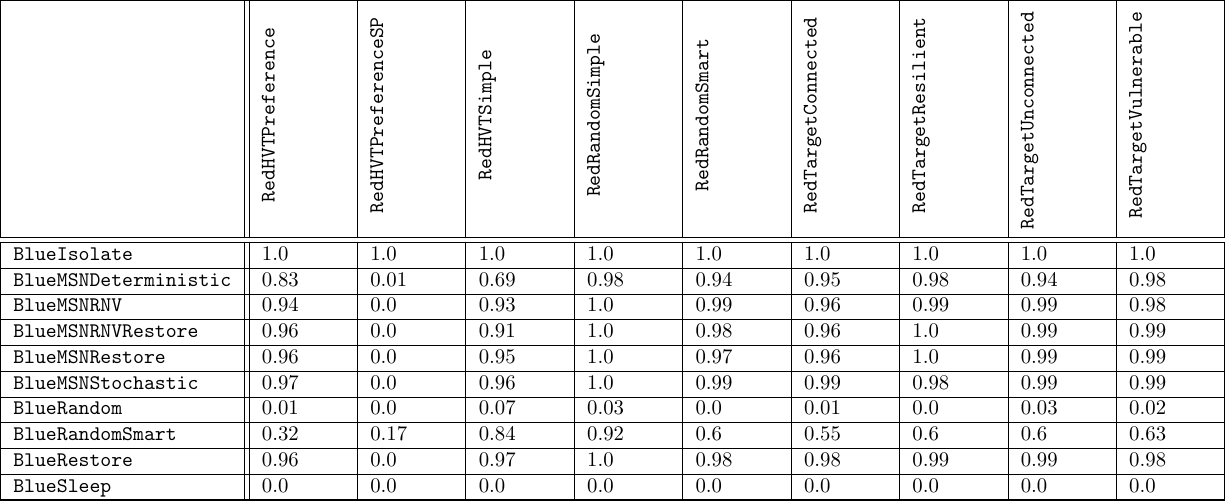}
    \caption{\texttt{OpticalCoreNetwork} Results for deterministic red agents ($\alpha=0.01$).}
    \label{table:OpticalCoreNetwork:results}
\end{table}

%% file: appendix/ntd_loss.tex
\section{The \mtrclong as a Loss Function} \label{app:ntd}

In our discussion of this work, we posit that the Network Transport Distance (NTD) holds potential for the direct optimization of predicted successor representations as well as evaluation.
While a comprehensive exploration lies beyond the scope of the current study, we here provide some preliminary experimental insights into this avenue.

In order to recast the \mtrc as a loss function, we took inspiration from an open-source PyTorch implementation of the Sinkhorn loss \cite{viehmann2019implementation} and replaced the \mtrc's enclosed linear program with Sinkhorn-Knopp iterations. This modification not only preserves computational efficiency but also ensures differentiability, thereby enabling its incorporation into gradient-based learning frameworks.

\medskip

\noindent Mathematically, the extension can be formalized as as follows:
\begin{equation}
    \ntdloss(P, Q, D, \lambda) = \frac{1}{\text{max}(D)}\inf_{\mu \in M(P, Q)} \left[ \int_{X \times X} d(i, j) \, d\mu(i, j) - \lambda H(\mu) \right],
    \label{eq:ntdloss}
\end{equation}
where:
\begin{itemize}
    \item \( \lambda > 0 \) is a user-specified regularization parameter.
    \item \( H(\mu) = -\sum_{i,j} \mu(i,j) \log \mu(i,j) \) is the entropy of the transport plan \( \mu \).
\end{itemize}

\noindent The optimal transport plan \( \mu^* \) is obtained via Sinkhorn-Knopp iterations, which iteratively update the transport plan \( \mu \) to satisfy the marginal constraints \( P \) and \( Q \):
\begin{equation}
    \mu^{(k+1)} = \text{diag}(u^{(k+1)}) \cdot K \cdot \text{diag}(v^{(k+1)}),
\end{equation}
where:
\begin{itemize}
    \item \( K_{ij} = \exp\left(-\frac{D[i, j]}{\lambda}\right) \) is a kernel matrix.
    \item The scaling factors \( u \) and \( v \) are updated as:
    \begin{equation}
        u^{(k+1)} = \frac{P}{Kv^{(k)}}, \quad v^{(k+1)} = \frac{Q}{K^T u^{(k+1)}}.
    \end{equation}
\end{itemize}

\begin{figure}[H]
    \centering
    \includegraphics[width=0.7\columnwidth]{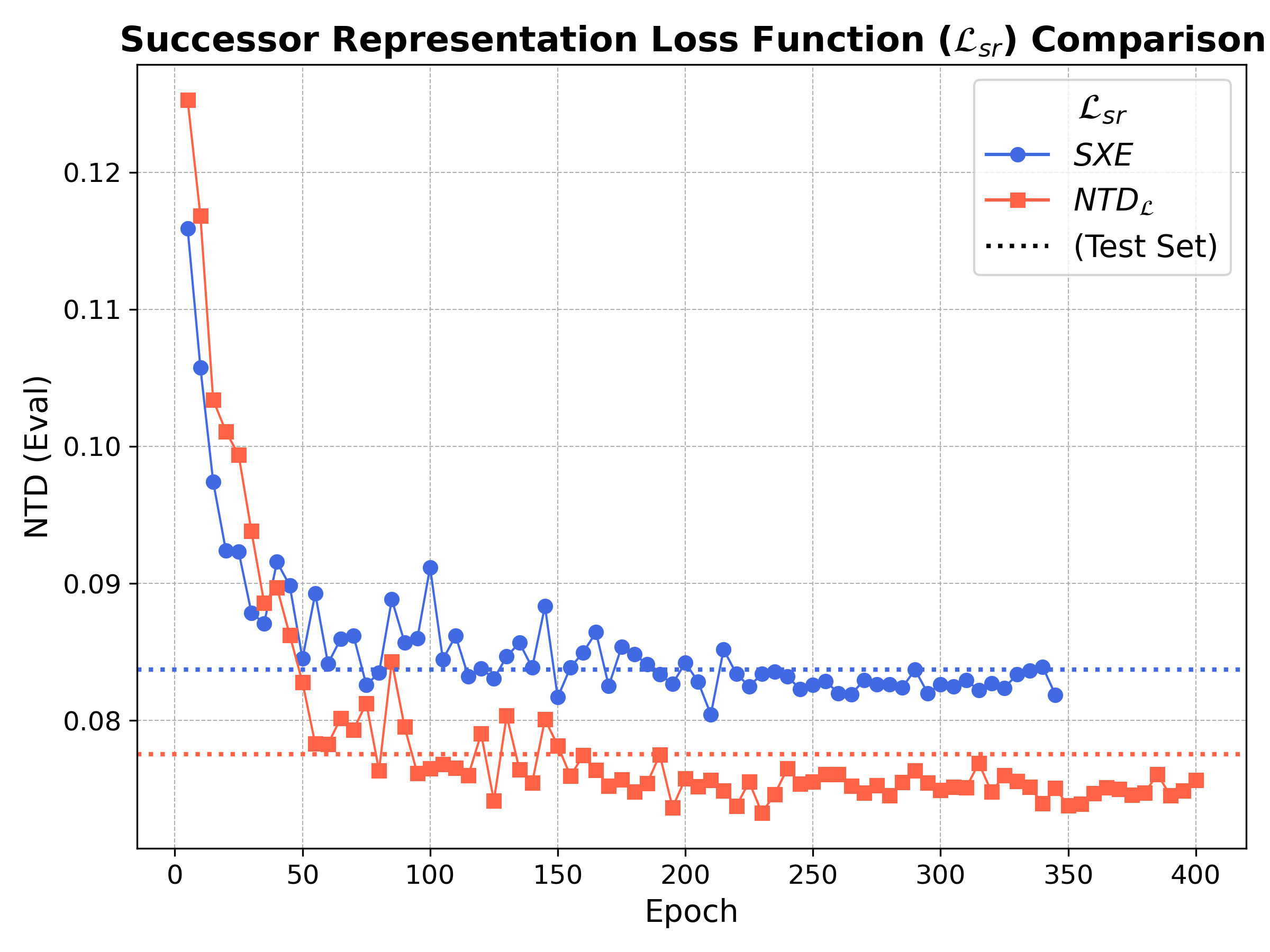}
    \caption{Evaluation and test set NTD scores for \gigo's predicted successor representations over \treeMixed when training \gigo only to optimize SR predictions using our \mtrc loss function, benchmarked against the soft label cross-entropy loss (SXE). Mean \mtrc scores are computed across each $\gamma \in \{0.5,0.95,0.999\}$. Hold-out test set NTD scores are overlaid with dotted lines.}
    \label{fig:loss_fn_comp_line}
\end{figure}

We then ran the preliminary test of comparing evaluation/test set \mtrc scores produced by this loss and the soft label cross-entropy loss when optimizing \gigo \emph{only} with respect to successor representation predictions (i.e. dropping $\Lhvt$ from \autoref{eq:total_loss} and instead having $\mathcal{L}_{total}=\Lsr$). Our results suggest that predicted successor representations are significantly improved when optimized with the $\ntdloss$ over the SXE (\autoref{fig:loss_fn_comp_line}).
